\documentclass[sigconf]{acmart}
\AtBeginDocument{%
	}

\copyrightyear{2022} 
\acmYear{2022} 
\setcopyright{rightsretained} 
\acmConference[MM '22]{Proceedings of the 30th ACM International
	Conference on Multimedia}{October 10--14, 2022}{Lisboa, Portugal}
\acmBooktitle{Proceedings of the 30th ACM International Conference on
	Multimedia (MM '22), October 10--14, 2022, Lisboa,
	Portugal}\acmDOI{10.1145/3503161.3547882}
\acmISBN{978-1-4503-9203-7/22/10}

\usepackage{graphicx}
\usepackage{amsmath}
\usepackage{booktabs}

\usepackage{setspace}
\usepackage{algorithm}
\usepackage{algorithmic}
\usepackage{multirow}
\usepackage{booktabs,tabularx, amsmath}
\usepackage{color}
\usepackage{float}
\usepackage{subfigure}
\usepackage{balance}
\usepackage{makecell}

\newfloat{figtab}{htb}{fgtb}
\makeatletter
\newcommand\figcaption{\def\@captype{figure}\caption}
\newcommand\tabcaption{\def\@captype{table}\caption}
\makeatother
\newcolumntype{Y}{>{\centering\arraybackslash}X} 

\usepackage[capitalize]{cleveref}
\crefname{section}{Sec.}{Secs.}
\Crefname{section}{Section}{Sections}
\Crefname{table}{Table}{Tables}
\crefname{table}{Tab.}{Tabs.}

\usepackage{etoolbox}
\makeatletter
\patchcmd{\maketitle}{\@copyrightpermission}{
	\begin{minipage}{0.3\columnwidth}
		\href{https://creativecommons.org/licenses/by/4.0/}{\includegraphics[width=0.90\textwidth]{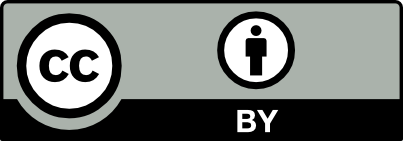}}
	\end{minipage}\hfill
	\begin{minipage}{0.7\columnwidth}
		\href{https://creativecommons.org/licenses/by/4.0/}{This work is licensed under a Creative Commons Attribution International 4.0 License.}
	\end{minipage}
	
	\vspace{5pt}
}{}{}

\makeatother

\begin{document}
	
	\title{TPSNet: Reverse Thinking of Thin Plate Splines for Arbitrary Shape Scene Text Representation}
	
	\author{Wei Wang}
	% \author{Yu Zhou} \authornote{Corresponding author.}
	% \author{Jiahao Lv}
	\email{wangwei3456@iie.ac.cn}
	\affiliation{%
		\institution{Institute of Information Engineering, Chinese Academy of Sciences}
		%   \city{Beijing}
		\country{}
	}
	\affiliation{%
		\institution{School of Cyber Security, University of Chinese Academy of Sciences}
		%   \city{Beijing}
		\country{}
	}
	
	\author{Yu Zhou} \authornote{Corresponding author.}
	\email{zhouyu@iie.ac.cn}
	\affiliation{%
		\institution{Institute of Information Engineering, Chinese Academy of Sciences}
		%   \city{Beijing}
		\country{}
	}
	\affiliation{%
		\institution{School of Cyber Security, University of Chinese Academy of Sciences}
		%   \city{Beijing}
		\country{}
	}
	
	\author{Jiahao Lv}
	\email{lvjiahao221@mails.ucas.ac.cn}
	\affiliation{%
		\institution{Institute of Information Engineering, Chinese Academy of Sciences}
		%   \city{Beijing}
		\country{}
	}
	\affiliation{%
		\institution{School of Cyber Security, University of Chinese Academy of Sciences}
		\country{}
	}
	\author{Dayan Wu}
	\email{wudayan@iie.ac.cn}
	\affiliation{%
		\institution{Institute of Information Engineering, Chinese Academy of Sciences}
		%   \city{Beijing}
		\country{}
	}
	\author{Guoqing Zhao}
	\email{guoqing.zhao02@msxf.com}
	\affiliation{%
		\institution{Mashang Consumer Finance Co., Ltd}
		%   \city{Beijing}
		\country{}
	}
	
	\author{Ning Jiang}
	\email{ning.jiang02@msxf.com}
	\affiliation{%
		\institution{Mashang Consumer Finance Co., Ltd}
		%   \city{Beijing}
		\country{}
	}
	\author{Weiping Wang}
	\email{wangweiping@iie.ac.cn}
	\affiliation{%
		\institution{Institute of Information Engineering, Chinese Academy of Sciences}
		%   \city{Beijing}
		\country{}
	}

	\renewcommand{\shortauthors}{Wei Wang et al.}
	
	\begin{abstract}
		The research focus of scene text detection and recognition has shifted to arbitrary shape text in recent years, where the text shape representation is a fundamental problem. An ideal representation should be compact, complete, efficient, and reusable for subsequent recognition in our opinion. However, previous representations have flaws in one or more aspects. Thin-Plate-Spline (TPS) transformation has achieved great success in scene text recognition. Inspired by this, we reversely think of its usage and sophisticatedly take TPS as an exquisite representation for arbitrary shape text representation. The TPS representation is compact, complete, and efficient. With the predicted TPS parameters, the detected text region can be directly rectified to a near-horizontal one to assist the subsequent recognition. To further exploit the potential of the TPS representation, the Border Alignment Loss is proposed. Based on these designs, we implement the text detector TPSNet, which can be extended to a text spotter conveniently. Extensive evaluation and ablation of several public benchmarks demonstrate the effectiveness and superiority of the proposed method for text representation and spotting. Particularly, TPSNet achieves the detection F-Measure improvement of 4.4\% (78.4\% vs. 74.0\%) on Art dataset and the end-to-end spotting F-Measure improvement of 5.0\% (78.5\% vs. 73.5\%) on Total-Text, which are large margins with no bells and whistles. 
	\end{abstract}

	
	
	
	\maketitle
	
	\section{Introduction}
	
	\begin{figure}[t]
		\centering
		\setlength{\abovecaptionskip}{5px}
		\subfigbottomskip=-4pt
		\subfigcapskip=-5pt
		\subfigure[Text rectification and representation based on TPS transform.]{\includegraphics[width=0.98\linewidth]{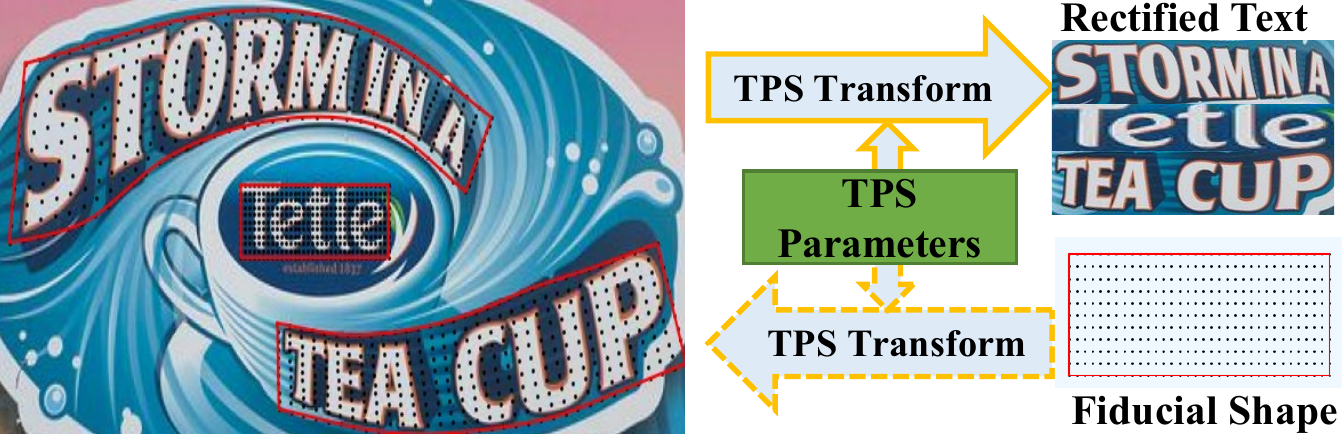}}
		\subfigure[TextRay  \cite{Wang2020textray}]{\includegraphics[width=0.323\linewidth]{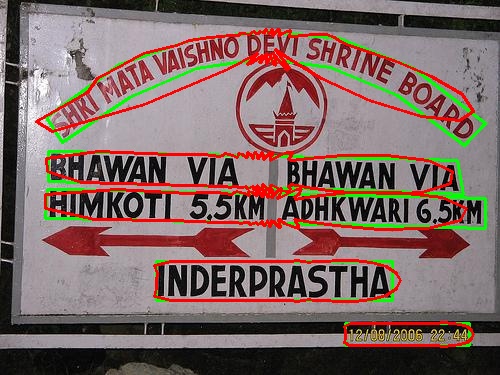}}
		\subfigure[FCE  \cite{zhu2021fourier}]{\includegraphics[width=0.323\linewidth]{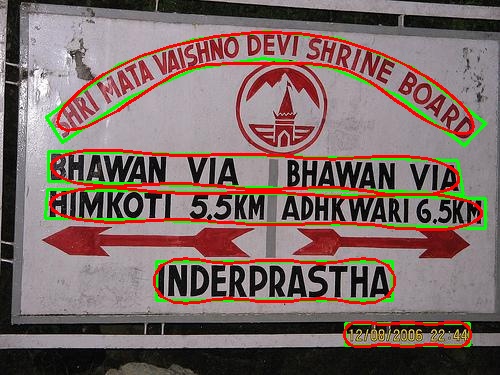}}
		\subfigure[TPS (Ours)]{\includegraphics[width=0.323\linewidth]{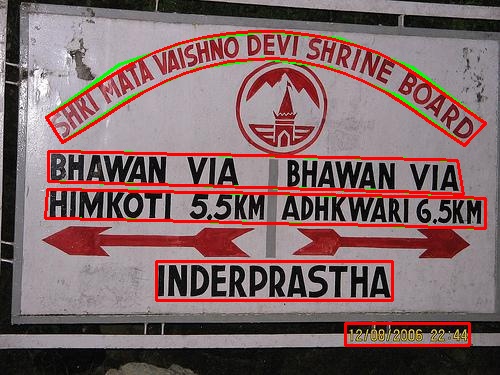}}
		\caption{Text shape representation methods. (a) shows that our proposed TPS representation can be seen as the reverse process of text rectification. Bottom is the text shape fitting results of (b) TextRay   \cite{Wang2020textray}, (c) FCE   \cite{zhu2021fourier} and (d) our proposed TPS. The red lines are the fitting curves and the green lines are the ground truth.
			The TextRay fails on highly-curved shapes and both of TextRay and FCE miss partial corner pixels on extreme aspect ratio cases.}
		\label{fig:intro}
		\vspace{-12px}
	\end{figure}
	
	Scene text detection and recognition have become increasing prevalent topics in computer vision, due to their various applications in document image analysis \cite{yang2017learning}, scene understanding \cite{zgy1}, autonomous driving \cite{zhu2017cascaded}, etc. Arbitrary shape text is one of the main challenges for detection and recognition, and lots of profound approaches have been proposed. For handling various text shapes, the fundamental problem is how to represent the shapes.
	
	Text shape representations can be classified into two types: segmentation-based methods and regression-based methods. Seg-mentation-based methods \cite{lyu2018masktextspotter,wang2019PSENet,qin2019towards,liao2020db} represent text regions with pixel-level classification masks that allow flexibility for arbitrary shapes, but they have drawbacks such as computationally intensive post-processing and the lack of noise resistance. The regression-based methods regress the text boundaries directly, making the prediction process much more straightforward. For horizontal and multi-oriented straight text, regressing the quadrilateral is sufficient to represent the text shape   \cite{zhou2017east, li2017towards,liao2018textboxes++, liu2018FOTS, he2018end, he2021most}. However, complex representations must be designed for arbitrary shape texts. Apart from directly increasing the number of points to represent text contours   \cite{wang2019ATRR, wang2019boundary,dai2021progressive, zhang2021adaptive}, some methods apply parametric curves to fit the text boundaries   \cite{Liu2020ABCNet,  Wang2020textray, zhu2021fourier,abcnetv2, gyh1},  resulting in tighter and smoother contour curves, but some of them are not suitable for the
	characteristics of text shapes as shown in Fig.~\ref{fig:intro}~(b) and (c).
	
	In our opinion, an ideal representation for arbitrary shape text should: (1) be compact and complete so that the presence of background pixels and the missing text pixels are as few as possible; 
	(2) be simple and efficient so that less time is consumed on detection;
	(3) facilitate following recognition, making curved texts be recognized simply and accurately.
	Although there are various representation methods for texts, most of them can not satisfy all the criteria. 
	
	To better represent arbitrary shape text, we propose a new representation: Thin-Plate-Spline (TPS) representation. TPS transformation  \cite{bookstein1989principal} is typically applied in scene text recognition for rectification   \cite{shi2016robust, shi2018aster, zhan2019esir,yang2019symmetry, shang2020character,qz1}, where the irregular text region is rectified to the horizontal regular region so that the classical simple methods like CRNN   \cite{crnn} can recognize it well. Though TPS is effective in scene text recognition,  it has not been directly applied for scene text representation to the best of our knowledge.
	If we think of the usage of TPS reversely, as shown in Fig. \ref{fig:intro} (a), it can be a novel and simple representation for arbitrary shape text.  
	When rectifying text for recognition, the TPS parameters are solved based on the corresponding control points, with which the source shape can be transformed into the target rectangular shape. Considering the reverse process of rectification, if getting the TPS parameters, every coordinate on the target rectangle can be transformed to the coordinate on the source arbitrary shape. Since the target rectangle can be fixed, the TPS parameter vector can be taken as an appropriate shape embedding and meet all the three criteria for ideal text representation. Firstly, the TPS representation takes the rectangle as the basic shape of text, and it is adaptive to the characteristics of large aspect ratios and right angles in the corners, as shown in Fig.~\ref{fig:intro}. Secondly, arbitrary shapes are encoded to the low-dimensional vectors that can be regressed directly, and decoding is also efficient. Thirdly, the TPS parameters can naturally rectify curved texts, leveraging the predicted shape for accurate recognition.
	
	Generally, the TPS parameters should be derived from the coordinates of control points. However, there is another problem that the boundary of the arbitrary shape text is casually annotated by sparse points without strict rules. Even the number of points is different. As a result, the ground truth of the control points lacks reliable definition, and the prediction for sparse points is not robust and hard to be optimized. So we choose to regress the TPS parameters directly and propose the Border Alignment Loss to supervise it, which abandons the hard point matching and exploits shape alignment to make the TPS parameters regression more accurate.
	
	\begin{figure*}[t]
		\centering
		\setlength{\abovecaptionskip}{2px}
		\subfigbottomskip=-3pt
		\subfigcapskip=-5pt
		\includegraphics[width=1.0\linewidth]{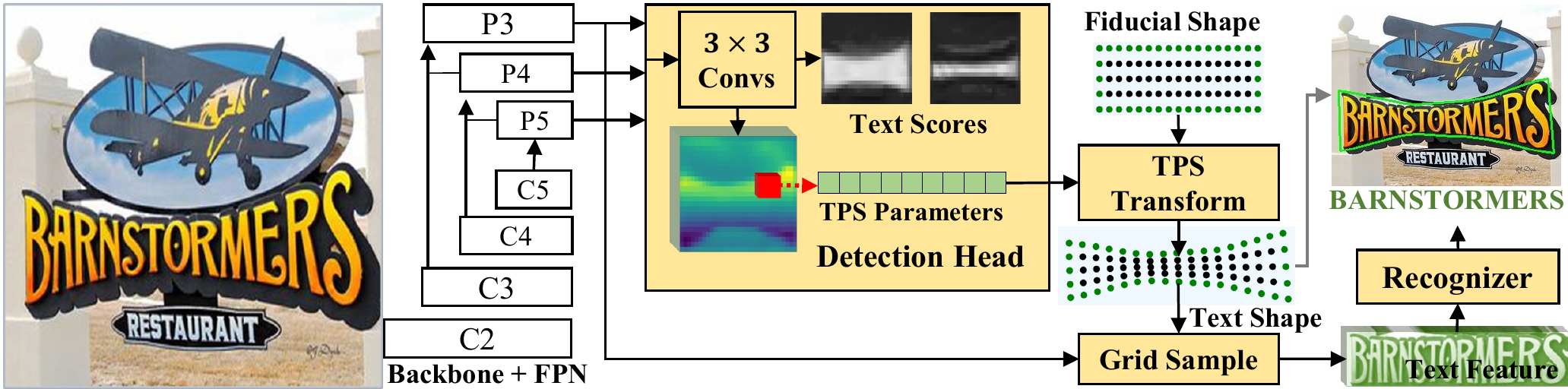}
		\caption{The architecture of the proposed TPSNet. The multi-level feature maps are extracted from the image by the backbone and
			FPN. Then the detection head with 3 x 3 convolutions classifies the text region and text center as the text scores and regresses
			the TPS parameter maps. The text scores will remove duplicated detection with NMS. The TPS parameters are
			transformed into text shapes with the pre-defined fiducial shape. The text boundary can be naturally obtained from the text
			shape, and the feature of arbitrary shape text is also rectified to an attached recognition head.
		}
		\label{fig:tpsnet}
		\vspace{-10px}
	\end{figure*}
	
	The contributions of this work are summarized as follows:
	\begin{itemize}
		\item An exquisite representation - TPS representation is first proposed for arbitrary shape text detection. The inspiration is from the TPS transformation used in text recognition extensively and sophisticated reverse thinking. It is compact, complete, efficient, and can be reused in recognition.
		\item To address the ambiguity of the boundary annotation and improve the supervision of the text shape, we design the Border Alignment Loss to exploit the potential of the TPS representation to obtain robust performance.
		\item TPSNet equipped with TPS representation and proposed loss function is presented, and it is extended to a text spotter with a simple recognizer.
		TPSNet is evaluated on several scene text detection and spotting benchmarks, and the performance is superior to previous counterparts.
	\end{itemize}
	
	\section{Related Works}
	\subsection{Segmentation-based Text Representation}
	As a special kind of object, the scene text needs to be represented appropriately for accurate detection. Segmentation mask is the common representation and has been widely used  \cite{deng2018pixellink, wang2019PSENet,tang2017scene,zhou2020crnet, lyu2018masktextspotter,liao2019masktextspotterv2, liao2020masktextspotterv3,liu2019maskTTD, liao2020db,ye2020textfusenet, xiao2020SD, liu2019CSE,liu2018MCN,tian2019SAE,yang2018inceptext, wang2019PAN,qxg2, xu2019textfield,xie2019SPCNet, zhu2021textmountain,cheng2019direct,xue2018border,rong2019unambiguous,guo2022units,wan2021self,qiao2020textperceptron,crafts,qxg3}. The mask can naturally represent arbitrary shape text, but it has the limitation of confusing different text instances
	and suffers from the lack of noise resistance and computationally intensive post-processing. Some methods represent the text with a set of text components  \cite{tian2016CTPN, shi2017SegLink, tang2019seglink++, zhang2020DRRG,ma2021relatext,feng2019textdragon, feng2021residual,cyd1}, which also belong to segmentation-based methods, but the units are text segments rather than pixels.
	
	\subsection{Regression-based Text Representation}
	Regressing the geometry of the text shape and position is another kind of representation. For the horizontal and multi-oriented straight text, a rectangle   \cite{zhang2016multi, zhou2017east, liao2017textboxes,he2017deep,liu2017deep, he2021most, wang2018ITN,he2017multi,qxg1} or quadrilateral  \cite{liao2018rotation, liao2018textboxes++, xue2018border, lyu2018corner,yuan2020follow} is sufficient. 
	For curved texts, the representation becomes complicated. TextSnake  \cite{long2018textsnake}and MSR  \cite{ xue2019msr} regress the distance to the text boundary, which is similar to segmentation-based methods. 
	Methods like  \cite{wang2019ATRR,wang2019boundary, dai2021progressive, zhang2021adaptive} directly regress the contour points as the text boundary. The points regression is more efficient than segmentation since there is no complicated post-processing. However, the position of every independent point on the boundary is not well defined when it comes to a variety of text shapes, making the points prediction not robust.
	
	The better choice is to represent text with the parameter curves, which abstract the text shape into a vector.
	TextRay   \cite{Wang2020textray} employs the Chebyshev polynomials under the polar coordinate system to approximate the boundary,
	but the distribution of sampling points under the polar system is not homogeneous on the boundary, making it hard to fit the long and highly curved texts. 
	FCENet  \cite{zhu2021fourier} adopts the Fourier series to fit the boundaries of highly curved texts. However, although the Fourier curve is an excellent fitter, it struggles to fit text at right angle corners with relatively fewer parameters, resulting in incomplete characters, as shown in Fig. \ref{fig:intro} (c).
	All the representations above have the limitation that they can not directly rectify the irregular shape text for subsequent recognition.
	ABCNet  \cite{Liu2020ABCNet, abcnetv2} formulates the long sides of the text with two Bezier curves to get compact border fitting and can rectify curved text by the interpolation grid, which is the only one that meets all three criteria among the previous methods.
	
	Our proposed TPS representation employs a fundamentally different fitting formula to provide another choice, which meets all three criteria and has more flexibility for the variations of the text shape. Furthermore, TPS representation is also superior to Bezier representation in the case of arbitrary-shape text with severe perspective distortion, which is frequently encountered in real applications though is scarce in public benchmark datasets.
	
	\section{Methodology}
	
	\subsection{Thin-Plate-Spline Representation}
	
	Text is a special kind of object. In standard cases such as document texts, the shape of a word or a text line is generally a (rotated) rectangle. In scene images, texts may be presented in various shapes containing distortion, while they are still deformed from the basic rectangle, mostly retaining the characteristics of right-angle corners and large aspect ratios. From the perspective of deformation, we try to establish a mapping between arbitrary text shapes to regular rectangles, thus enabling arbitrary shape text representations.
	
	\begin{figure}
		\vspace{-0.2cm} 
		\centering
		\setlength{\abovecaptionskip}{5px}
		\subfigbottomskip=-3pt
		\subfigcapskip=-5pt
		\subfigure[The TPS transformation process]{\includegraphics[width=0.98\linewidth]{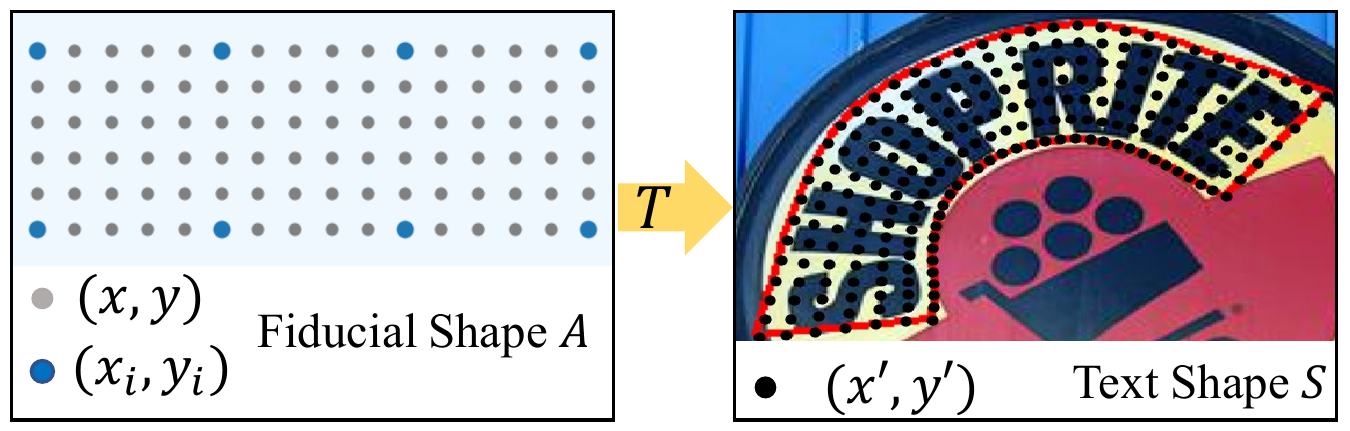}}
		\subfigure[Edge]{\includegraphics[width=0.32\linewidth]{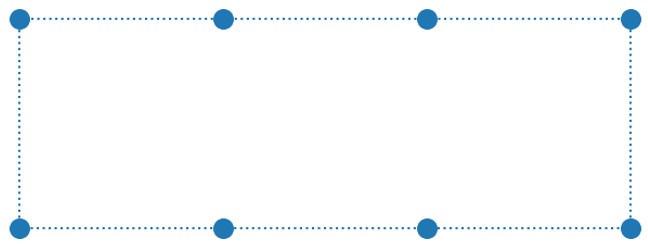}}
		\subfigure[Cross]{\includegraphics[width=0.32\linewidth]{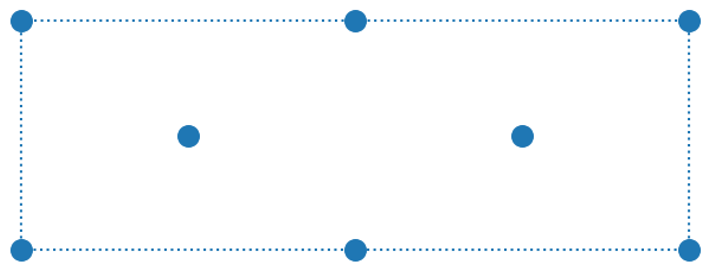}}
		\subfigure[Center]{\includegraphics[width=0.32\linewidth]{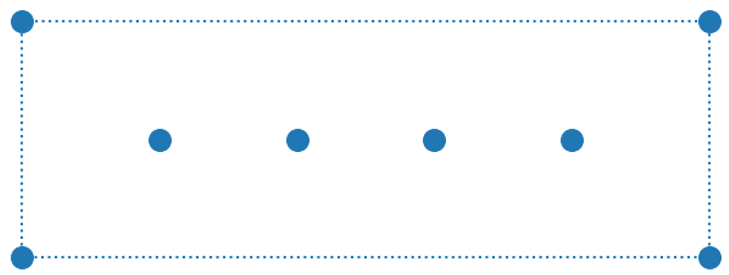}}
		\caption{(a) Illustrations of the TPS transformation process. (b), (c) and (d) are three different distributions of the fiducial points $(x_i,y_i)$, which define the basis function $\phi(x,y)$.}
		\label{fig:fid}
		\vspace{-8px}
	\end{figure}
	
	TPS \cite{bookstein1989principal} has been widely used as the non-rigid transformation model in image alignment and shape matching. We apply TPS as the basic model to implement the deformation from an arbitrary shape  to a regular rectangle, which requires obtaining the correspondence of every coordinate from the rectangle to an arbitrary shape . We formulate the "target" shape as the rectangle $A$ and the "source" text shape as $S$, and the rectangle $A$ is also called fiducial shape. According to TPS   \cite{bookstein1989principal}, the corresponding point of $(x,y) \in A$ on $S$ can be calculated by
	\begin{equation}
	\begin{array}{l}
	x' = \Phi_x(x,y) = c_x + a_{1_x} x + a_{2_x} y + \sum_{i=1}^{k}w_{i_x} r(d_i) \\
	y' = \Phi_y(x,y) = c_y + a_{1_y} x + a_{2_y} y + \sum_{i=1}^{k}w_{i_y} r(d_i) \\
	\end{array}
	\label{equ:1}
	\end{equation}
	%其中 r是径向基函数
	where $r$ is the radial basis function
	\begin{equation}
	r(d) = \left\{
	\begin{array}{l}
	0, \quad \quad \quad if \ d = 0\\
	d^2ln d, \quad otherwise
	\end{array}
	\right.
	\end{equation}
	$d_i=||(x,y)-(x_i,y_i)||_2$ is the distance from $(x,y)$ to $(x_i,y_i)$. \{($x_1,y_1$), $(x_2,y_2), ... , (x_k,y_k)$\} are the fixed points on the fiducial shape A, called fiducial points, and $k$ is the number of fiducial points. Given fiducial points, the basis function is defined as 
	\begin{equation}
	\mathbf{\phi}(x,y) = \left [\begin{array}{c}
	1 \quad x \quad y \quad r(d_1) \quad ... \quad r(d_k)
	\end{array} \right] ^ \mathrm{ T }
	\end{equation}
	then the TPS transform function is determined by the parameters \begin{equation}
	\mathbf{T} = \left [
	\begin{array}{cccccc}
	c_x & a_{1_x} & a_{2_x}  & w_{1_x} & ... &  w_{k_x}\\
	c_y & a_{1_y} & a_{2_y} &  w_{1_y} & ... &  w_{k_y}
	\end{array}
	\right ]
	\end{equation} with the shape of $2\times(k+3)$.

	With the TPS transform function, 
	the grids $(x,y)$ on $A$ can be  transformed into the corresponding points on $S$, where the text boundaries are naturally obtained, 
	as shown in Fig.~\ref{fig:tpsnet}. Note that, the grids on $A$ are  predefined, so the $\phi(x,y)$ is also calculated in advance. The TPS parameters $\mathbf{T}$ can be decoded quickly to text shape with a matrix multiplication 
	\begin{equation}
	(x',y')^T = \textbf{T} \phi(x,y)
	\label{equ:5}
	\end{equation}

	Any set of points not lying on the same line can be taken as the fiducial points to define a basis function $\phi(x,y)$, which determines the fitting ability of the TPS parameters. For simplicity and efficiency, the number of fiducial points $k$ is set to 8, thus the dimension of TPS parameters $\mathbf{T}$ is 22, which is enough to fit nearly all the text shapes. There are different options for the distribution of the fiducial points, as shown in Fig.~\ref{fig:fid}. Except for the four corner points that should be fixed to determine the base location, the other fiducial points can locate not only on the edges like (b) but also inside the shape like (c) cross and (d) center. The corresponding points on text shape $S$ 
	will distribute in the same relative position. This is different from Bezier \cite{Liu2020ABCNet}, which uses two sets of control points to fit two edges separately. 
	With more points located along the width, the longer curved text can be fitted, while with more points on the edge, the text boundary will be more accurate. Visualizations of fitting results with different distributions can be found in Appendix~\ref{sec:a2}. The ablation study is conducted to decide the optimal distribution.
	
	We emphasize that, unlike previous detection or rectification methods, we do not attempt to predict control points on the text shape $S$ because the locations of control points on arbitrary shapes are not well-defined, 
	and we directly regress the TPS parameters $T$ with the neural network instead.

	To verify the fitting ability of TPS,
	we solve the TPS parameters $\mathbf{T}$ with the standard least square method from equation (\ref{equ:5}).
	The visualization of the shape fitting is shown in the Fig. \ref{fig:intro}, and the quantitative evaluation can be found in Appendix~\ref{sec:a1}. The TextRay \cite{Wang2020textray} fails on highly-curved shapes, and both TextRay \cite{Wang2020textray} and Fourier \cite{zhu2021fourier} fail on extreme aspect ratio cases and miss the text corners, and they can not rectify irregular text for subsequent recognition either.
	
	\begin{figure}
		\centering
		\setlength{\abovecaptionskip}{2px}
		\subfigbottomskip=-3pt
		\subfigcapskip=-5pt
		\includegraphics[width=0.98\linewidth]{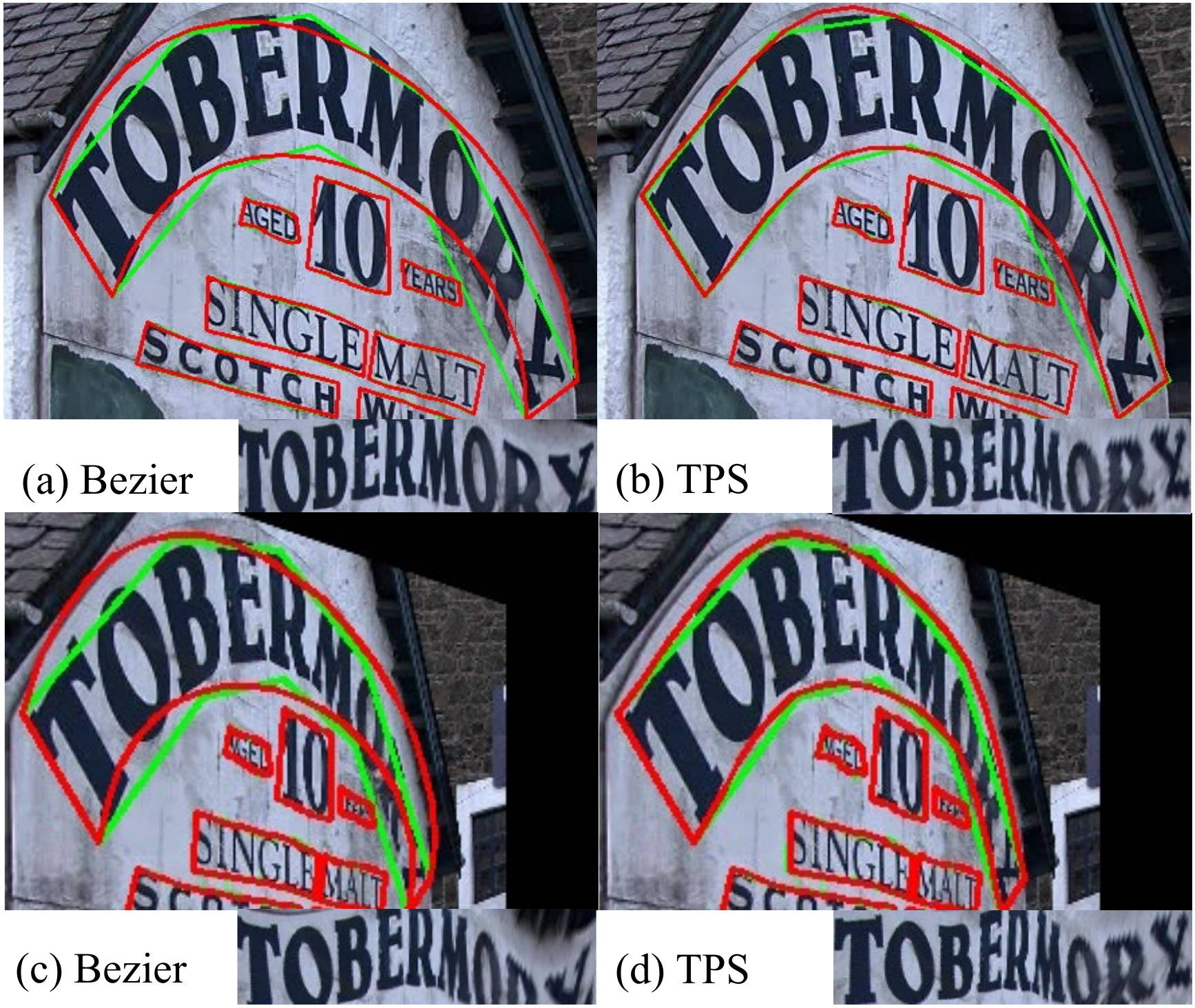}
		\caption{Illustrations of the text shape fitting and rectification with Bezier \cite{Liu2020ABCNet} and TPS. (a),(b) are the original images, (c) and (d) are the perspective images. The green line is the ground truth, and the red line is the fitting result.}
		\label{fig:tps_vs_bezier}
		\vspace{-12px}
	\end{figure}
	
	The Bezier \cite{Liu2020ABCNet, abcnetv2} representation is the closest one to TPS, but it employs a different fitting formula:
	\begin{equation}
	(x',y')^T = \left [
	\begin{aligned}
	w_{0_x} & ... &  w_{n_x}\\
	w_{0_y} & ... &  w_{n_y}
	\end{aligned}
	\right]
	[B_{0,n}(t) \quad ... \quad B_{n,n}(t)] ^T
	\end{equation}
	where $(w_{i_x}, w_{i_y})$ is the $i_{th}$ control points and $B_{i,n}(t)$ represents the Bernstein basis polynomials \cite{Liu2020ABCNet}. Our TPS representation Equ~(\ref{equ:5}) can be rewritten as:
	\begin{equation}
	\begin{aligned}
	(x',y')^T = & \left [
	\begin{array}{ccc}
	c_x & a_{1_x} & a_{2_x}  \\
	c_y & a_{1_y} & a_{2_y} 
	\end{array}
	\right ]
	[1 \quad x \quad y] ^T + \\
	&\left [
	\begin{aligned}
	w_{1_x} & ... &  w_{k_x}\\
	w_{1_y} & ... &  w_{k_y}
	\end{aligned}
	\right] [r(d_1) \quad ... \quad r(d_k)]^T
	\end{aligned}
	\end{equation}
	The Bezier formula is similar to the second part of the TPS formula, which reveals the local shapes. However, it misses the first part, which represents the global affine transformation. In addition, the basis function of Bezier $B_{i,n}(t)$ is one-dimensional function, while that of TPS $\phi(x,y)$ is two-dimensional. These differences allow our proposed TPS representation to handle more variations of text shapes. As shown in Fig.~\ref{fig:tps_vs_bezier}, Bezier and TPS with the same number of control points are used to fit the text in a perspective image. As (a) and (b) show, the TPS fits better than Bezier. In (c) and (d), when the perspective degree of the image is enlarged, 
	Bezier fails to fit this perspective curved text, and the rectification of the text is incomplete while our TPS still works well.

	\subsection{Border Alignment Loss}
	With the TPS transformation, the geometric shape is abstracted to the TPS parameter $\mathbf{T}$, and coordinates in the shape space are determined by all elements of  $\mathbf{T}$ in parameter space. Directly optimizing $\mathbf{T}$ in parameter space with distance loss like mean absolute error treats each element independently and neglects the intra-parameter correlations \cite{Wang2020textray}. To keep the inherent properties of the TPS transformation,
	$\mathbf{T}$ should be decoded into the shape space first, and the loss can be calculated as the distance of boundary point pairs between the decoded shape and its ground truth.
	
\begin{figure}[t]
		\vspace{-0.15cm} 
		\centering
		\setlength{\abovecaptionskip}{5px}
		\subfigbottomskip=-3pt
		\subfigcapskip=-5pt
		\subfigure[Polygon annotation]{\includegraphics[width=0.47\linewidth]{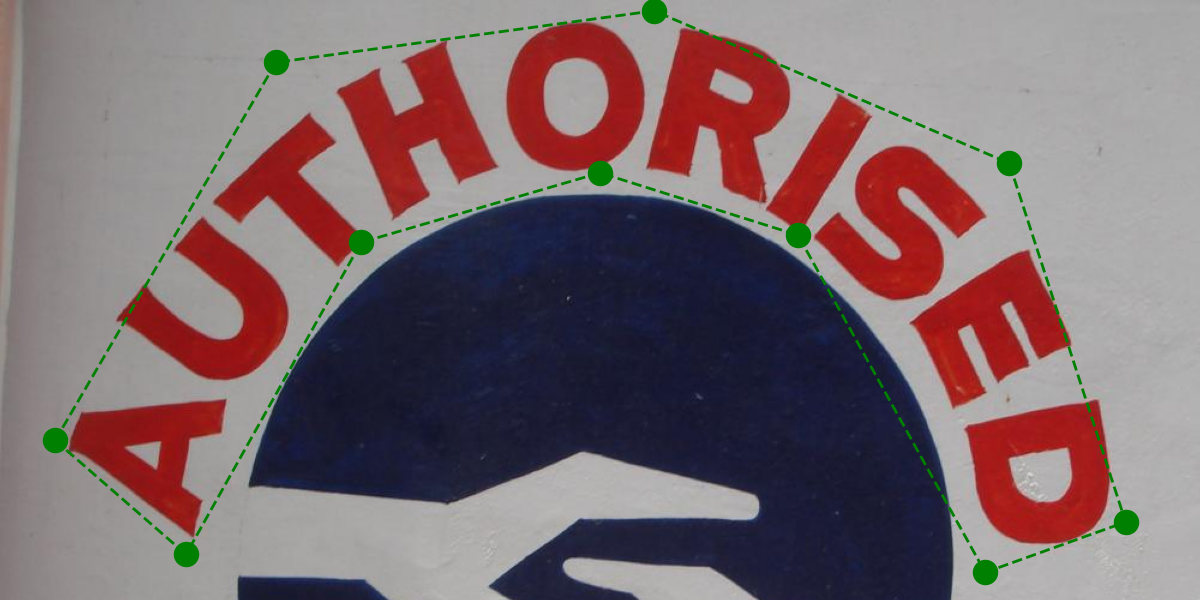}}
		\subfigure[Points regression]{\includegraphics[width=0.47\linewidth]{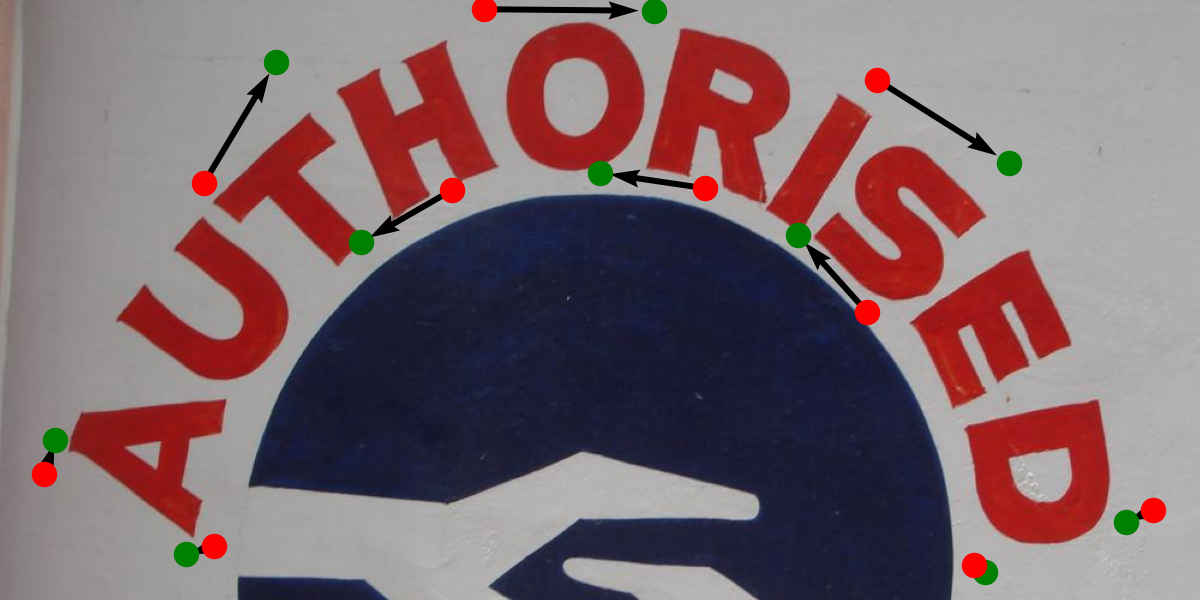}}
		\subfigure[Border Alignment Loss]{\includegraphics[width=0.95\linewidth]{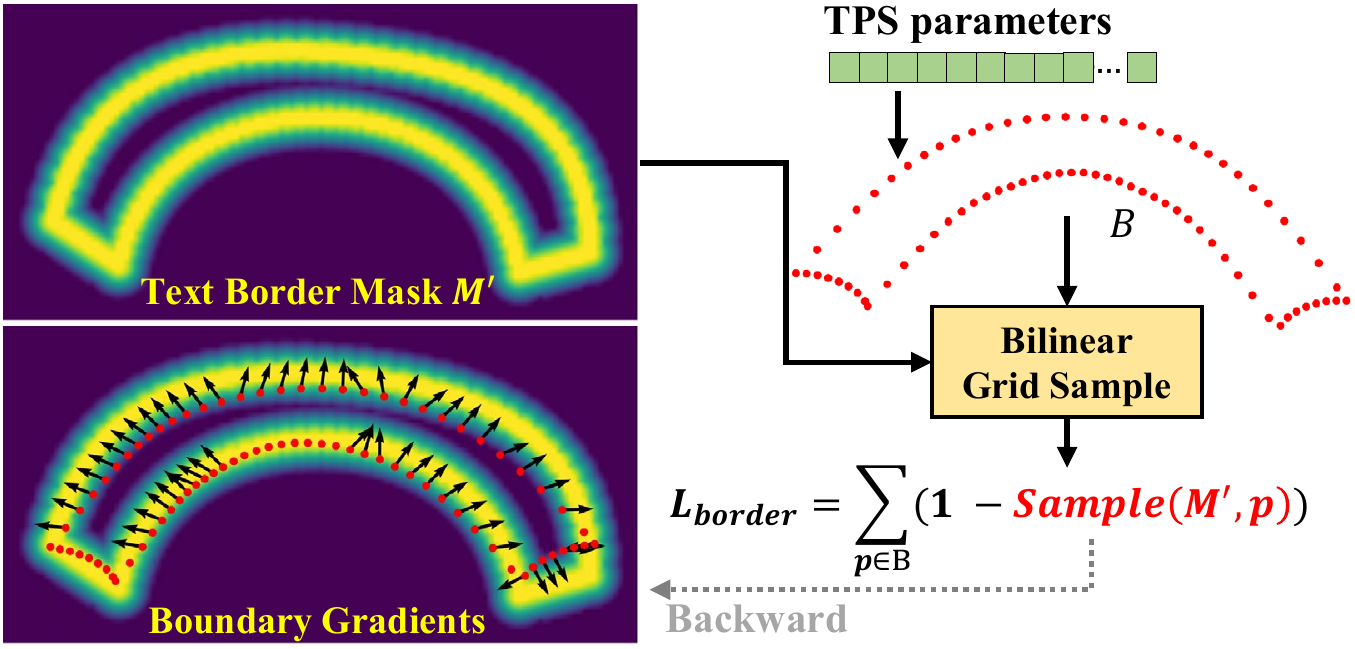}}
		\caption{ Illustrations of the Border Alignment Loss. (a) The text boundary annotation. (b) The points regression method. Red points are the regressed results, and green points are the ground truth. Although the regressed points have already located on the text boundary, some of them are still far from the ground truth (black arrows). (c) The proposed Border Alignment Loss.}
		\label{fig:loss}
		\vspace{-15px}
	\end{figure}

	As shown in Fig.~\ref{fig:loss}~(a), the text boundaries are annotated with sparse points. These limited number of points roughly describe the boundary, and because of the lack of strict definition, the positions of these points are not unique. The number of the annotation point is even different across datasets. For example, CTW1500 uses 14 points while Total-Text uses ten or less. In other words, the ground truth of the boundary points is ambiguous and noisy. Directly regressing points to the ground truth points will bring noise and ambiguity to the network optimization. As shown in Fig.~\ref{fig:loss}~(b), the prediction (red points) have already located on the text boundary and can describe this text instance well, but some points are still far from the noisy ground truth (green points), the undesired loss will be backward to the network, disrupting its convergence.
	
	To address this problem, we propose the Border Alignment Loss (BA-Loss). As shown in Fig.~\ref{fig:loss}~(c),
	Firstly, the noisy polygon annotation is smoothed with cubic spline interpolation\cite{mckinley1998cubic}, and than
	the text border mask $M$  is generated as follows: 
	\begin{equation}
	M_{xy} = \left\{
	\begin{aligned}
	&0, &\frac{d_{xy}}{s} \geq t_b\\
	&1-\frac{d_{xy}}{s\cdot t_b}, &\frac{d_{xy}}{s} < t_b
	\end{aligned}
	\right.
	\end{equation}
	where $M_{xy}$ is the value at point $(x,y)$ on text border mask, $d_{xy}$ is the minimum distance from $(x,y)$ to the smoothed boundary in Fig.~\ref{fig:loss}~(a), $s$ denotes the height of this text instance, and $t_b$ is the distance threshold, set to 0.6 empirically. This border mask is similar to the threshold map in DBNet \cite{liao2020db}, but here it is used for regression supervision. To further alleviate the noise of annotation, the text border mask is relaxed by:
	\begin{equation}
	M_{xy}^{'} = \left\{
	\begin{aligned}
	&1, &M_{xy} \geq t_r\\
	&\frac{M_{xy}}{t_r}, &M_{xy} < t_r
	\end{aligned}
	\right.
	\end{equation}
	where $M^{'}_{xy}$ is the relaxed mask value, $t_r$ is the relaxation threshold.
	Finally the text border mask is shown as Fig.~\ref{fig:loss}~(c).  Within a range on the boundary line, the pixel values are all equal to 1.0, and outside this range, the pixel values decrease to 0 as the distance increases.
	
	Then the continuous boundary points $B$ are decoded from the regressed $\mathbf{T}$, and the values at these points are sampled from the text border mask $M^{'}$ with bilinear grid sample. The loss for every text instance is calculated with:
	\begin{equation}
	L_{BA}(B) = \frac{1}{|B|}\sum_{p\in B}(1 - Sample(M^{'},p))
	\end{equation}
	where $Sample$ means the grid sample function with bilinear interpolation. $L_{BA}$ pulls all the predicted points near the border of the text, as the gradients shown in Fig.~\ref{fig:loss}~(c). For the points that have already located on the border, no gradient is generated. With $L_{BA}$, the predicted border is aligned to the annotation boundary with more tolerance to the noise and ambiguity.
	
	Because $L_{BA}$ only works when the predicted boundaries are not far from the text, so another corner point loss $L_{cor}$ is necessary to restrict basic location. $L_{cor}$ is calculated as the distance for 4 corner points to their ground truth:
	\begin{equation}
	L_{cor}(B) = \frac{1}{4}\sum_{p\in B_{corner}}||p-p_{gt}||_2
	\end{equation}
	The annotations of the corner are relatively robust, so $L_{cor}$ will not cause ambiguity.
	
	The total regression loss is:
	\begin{equation}
	L_{reg} = \frac{1}{N}\sum_{t=0}^N \frac{1}{|\Omega_t|}(L_{BA}(B_t) + L_{cor}(B_t))
	\end{equation}
	where $|\Omega_t|$ denotes the area of text instance $t$, $B_t$ means the boundary of $t$, and $N$ is the total number of text instances.

	\subsection{TPSNet}
	\subsubsection{Text Detection}  
	Following previous regression-based text detection network    \cite{zhou2017east, Wang2020textray, zhu2021fourier}, we adopt a compact one-stage fully-convolutional framework. As shown in Fig.\ref{fig:tpsnet}, for an input image, multi-scale features are extracted with the backbone and Feature Pyramid Network (FPN) \cite{lin2017FPN}. The detection head takes each level feature as input and uses convolution layers to predict text scores and TPS parameters maps. Following Textsnake \cite{long2018textsnake}, the text scores consist of the per-pixel masks of Text Region (TR) and Text Center (TC) classifications, and these two masks are multiplied as the confidence of the detection for every position. For every feature bin located inside of TR, the TPS parameters vector $\mathbf{T}$ is regressed for the corresponding text instance. Then $\mathbf{T}$ is decoded to text shape grids by TPS transform with Equ~(\ref{equ:5}), where the fiducial shape is a pre-defined grid. From the text shape, the text boundary can be obtained directly, and it is also convenient to rectify the text by sampling on the input image. Duplicated detection results are removed with Non-Maximum Suppression (NMS).
	
	For the TC classification, the Text Center Region(TCR) in previous methods are replaced with our proposed Gaussian Text Center(GTC), details can be found in Appendix~\ref{sec:b2}.
	
	The optimization objectives of the classification branch and regression branch are $L_{cls}$ and $L_{reg}$ respectively, and the TPSNet for detection is optimized by:
	\begin{equation}
	L = L_{cls} + L_{reg}
	\end{equation}
	The classification loss consists of the Text Region loss $L_{TR}$ and Text Center Region loss $L_{TC}$:
	\begin{equation}
	L_{cls} = L_{TR} + L_{TC}
	\end{equation}
	Both $L_{TR}$ and $L_{TC}$ are cross entropy losses. To solve the sample imbalance problem, OHEM \cite{zhu2021fourier} is adopted for $L_{TR}$.
	%with the ratio between negative and positive samples being 3 : 1.
	
	\subsubsection{Text Spotting}
	The detection model can be extended to an end-to-end text spotter with ease because it is convenient for the TPS representation to align the feature of the irregular shape text. The feature vector at each text shape grid point on the feature map is obtained with grid sampling as Spatial Transformer Network (STN) \cite{jaderberg2015spatial} to compose the text feature.
	
	Any recognition model can be applied for the recognition. For simplicity, we take the same recognition module as ABCNetV2 \cite{abcnetv2}, which consists of 6 convolutional layers, one bidirectional LSTM layer, and an attention-based decoder.  For end-to-end training, the whole loss function is 
	\begin{equation}
	L = L_{cls} + L_{reg} + L_{rec}
	\end{equation}
	where $L_{rec}$ is the Cross Entropy Loss for the recognition as in ABCNetV2 \cite{abcnetv2}.
	
	\section{Experiments}
	In this section, we evaluate our proposed TPSNet by CTW1500, Total-Text,  ICDAR2015 and Art datasets to validate its effectiveness. We first conduct some ablation studies to demonstrate the advantages of proposed designs and the setting of hyper-parameters. Then we compare the detection performance of our model with previous state-of-the-art methods. Finally, the spotting performance on arbitrary shape scene text datasets are evaluated.
	
	\subsection{Datasets}
	\noindent \textbf{Total-Text}  \cite{ch2017total} includes curved, horizontal, and multi-oriented text. It consists of 1, 255 training images and 300 test images. All text annotations are word-level. Text areas are annotated with polygons.\\
	\textbf{SUCT-CTW1500} \cite{yuan2019ctw} is a dataset for the curved text. It contains 1,000 training images and 500 test images. Text is represented by polygons with 14 points at text-line level.\\
	\textbf{ICDAR2015}  \cite{karatzas2015icdar2015} is a multi-oriented text detection dataset only for English, which includes 1000 training images and 500 testing images. The text regions are annotated with quadrilaterals.\\
	\textbf{ArT} \cite{chng2019icdar2019} is a large-scale multi-lingual arbitrary shape
	scene text detection dataset and is one of most complex datasets. It includes 5,603 training images and 4,563 testing images. The text regions are annotated by the polygons with adaptive number of key points.\\
	\textbf{SynthText-150K}  \cite{Liu2020ABCNet} is the synthetic datasets generated based on the method from  \cite{gupta2016synthetic}, and it includes nearly 150k images that contains straight and curved texts. 
	
	\subsection{Implementation Details}
	We implement our TPSNet based on MMOCR   \cite{kuang2021mmocr} with Pytorch   \cite{paszke2019pytorch} library. The backbone is ResNet50 pretrained on ImageNet with DCN in stage 2, 3 and 4, followed by FPN. Feature maps of P3, P4 and P5 are used in classification and regression branch, and P2, P3, P4 are used for recognition. 4 convolutional layers of $3\times3$ are applied for the text region and text center classification and the TPS regression.
	Text instances are assigned into different feature maps according to its scale ratio (instance scale/image scale).
	%, and the ranges are [0, 0.25], [0.2, 0.65] and [0.55, 1.0] for P3, P4 and P5 respectively.
	
	The training images are resized to $800\times800$, and data augmentation strategies are applied, including ColorJitter, RandomCrop, and RandomRotate. The training batch size is set to 8. Stochastic gradient descent (SGD) is adopted as an optimizer with a weight decay of 0.001 and a momentum of 0.9. For the detection-only training, the learning rate is initialized at 0.001 and adjusted by the Poly policy with a power of 0.9. The TPSNet is pretrained on Synthtext-150K datasets for 150k iteration, and the numbers of the finetune epochs are 250, 500, 500, and 200 for Total-Text, CTW1500, ICDAR2015, and Art separately. For end-to-end text spotter, the training schedule is set totally according to ABCNet V2 \cite{abcnetv2}, it is first pretrained on Synthtext-150K and MLT17 (English-only) for 260k iteration, and then finetuned 10k iterations for Total-Text and 80k for CTW1500.
	
	In the test stage, the short sides of the test image are set to 736, 736, 1080, 1600 for Total-Text, CTW1500, ICDAR2015, and ArT, while the long sides are resized to keep the original aspect ratio. All experiments are conducted on a single NVIDIA RTX3090 GPU.
	
	\begin{table}[t]
		\setlength{\abovecaptionskip}{0cm}  %段前
		\small
		\caption{Ablation study about representations and supervisions on Total-Text. ``Rep'' means Representation, including directly regressing points or TPS parameters. $L(B)$ means matching predicted boundary points to ground truth points as Fig.~\ref{fig:loss}~(b). TCR means the text center region, GTC means the gaussian text center. $L_{cor}$, $L_{BA}$ and Border Relax are our proposed losses in Section 3.2.}
		\centering
		\renewcommand{\arraystretch}{0.8}
		\begin{tabularx}{\linewidth}{@{}lccccccY@{}}
			\toprule
			Rep                  & $L(B)$       & TCR        & GTC        & $L_{cor}$  & $L_{BA}$   & Border Relax     & H    \\ \midrule
			Points               & \checkmark   &  \checkmark &            &             &  &  & 81.3 \\ \midrule
			\multirow{5}{*}{TPS} & \checkmark   &   \checkmark &            &             &  & & 83.5 \\
			& \checkmark   &             &\checkmark   &             &  &  & 84.7 \\
			&              &             & \checkmark  & \checkmark   &   &   & 56.7 \\
			&              &             & \checkmark &  \checkmark  & \checkmark   &         & 85.8     \\
			&              &             & \checkmark & \checkmark  & \checkmark & \checkmark & \textbf{86.6} \\ \bottomrule
		\end{tabularx}
		\label{tab:ablaloss}
		\vspace{-15px}
	\end{table}

	\subsection{Ablation Study}
	% \subsubsection{TPS Representations and $L_{BA}$}
	The ablation study is conducted  on the Total-Text dataset without pretraining.
	%\subsubsection{Regression Target and Supervision}
	As shown in Table~\ref{tab:ablaloss}, to verify the advantage of the TPS representation, we set a baseline model that directly regresses the boundary points and uses the simple boundary distance loss $L(B)$ as Fig.~\ref{fig:loss}~(b) and TCR for classification.
	%to verify the advantage of the TPS representation, it is compared with directly regressing boundary points, and both of them employ the simple boundary distance loss $L(B)$ as Fig.~\ref{fig:loss}~(b) and . 
	Compared with the points representation, TPS representation brings 2.2\% improvements on Hmean, proving abstracting geometric shapes to TPS parameter space can represent text better. The Gaussian Text Center (GTC) performs 1.2\% better than the Text Center Region (TCR). The boundary distance loss $L(B)$ is not a good choice to optimize the TPS regression as we have discussed above, so we replace it with our proposed Border Alignment Loss. The corner loss $L_{cor}$ alone can only locate the basic position of text but not fit curved texts, while $L_{BA}$ makes up this part, they work together to bring 1.1\% improvements on Hmean, and the border relaxation brings another 0.8\% improvement. The relax threshold $t_r$ is set to 0.8, of which the ablation is present in Appendix~\ref{sec:b1}. Note that the Border Assignment Loss leverage the continuity of TPS representation that dense boundary points can be decoded from it, so $L_{BA}$ can not apply to the direct points regression. In other words, the TPS representation brings totally the 5.3\% improvement on Hmean.
	The ablation study about the distribution of the fiducial points is present in Appendix~\ref{sec:a2}, the Cross is selected as the default setting.

	\subsection{
		%Comparisons with Detection Only Methods
		Detection Evaluation}
	
	\begin{table}[ht]
		\setlength{\abovecaptionskip}{0cm}  %段前
		\small
		\caption{Comparison on highly curved text subset of CTW1500 test set with variations on the degree of perspective. The Performances are Hmeans under IOU constant at 0.7. The perspective augmentation is applied to training set, and the test set is perspective with an angle at at $0^\circ$ (original), $45^\circ$ and $70^\circ$ for evaluation.}
		\centering
		\renewcommand{\arraystretch}{0.8}
		\begin{tabularx}{\linewidth}{@{}lYYYY@{}}
			\toprule
			\multirow{2}{*}{Reps} & \multirow{2}{*}{\begin{tabular}[c]{@{}l@{}}Perspective\\ Augmentation\end{tabular}} & \multicolumn{3}{c}{H(IOU@0.7)} \\ \cmidrule(l){3-5} 
			&  & $0^\circ$ & $45^\circ$ & $70^\circ$ \\ \midrule
			\multirow{2}{*}{Bezier} &    & 71.6 &   64.8      &  54.7          \\
			& \checkmark   & 71.3 (-0.3) &  67.6(+2.8)   &  60.0 (+5.3)       \\ \midrule
			\multirow{2}{*}{TPS}    &           &      72.1   &   66.2       &   56.2       \\
			& \checkmark &  72.3(+0.2)   &   71.2(+4.0)   &  65.2 (+9.0)   \\ \bottomrule
		\end{tabularx}
		\label{tab:bezier_vs_tps}
		\vspace{-12px}
	\end{table}
	
	% TODO 添加更多解释
	
	\begin{figure}[t]
		\setlength{\abovecaptionskip}{2px}
		\centering
		\includegraphics[width=0.9\linewidth]{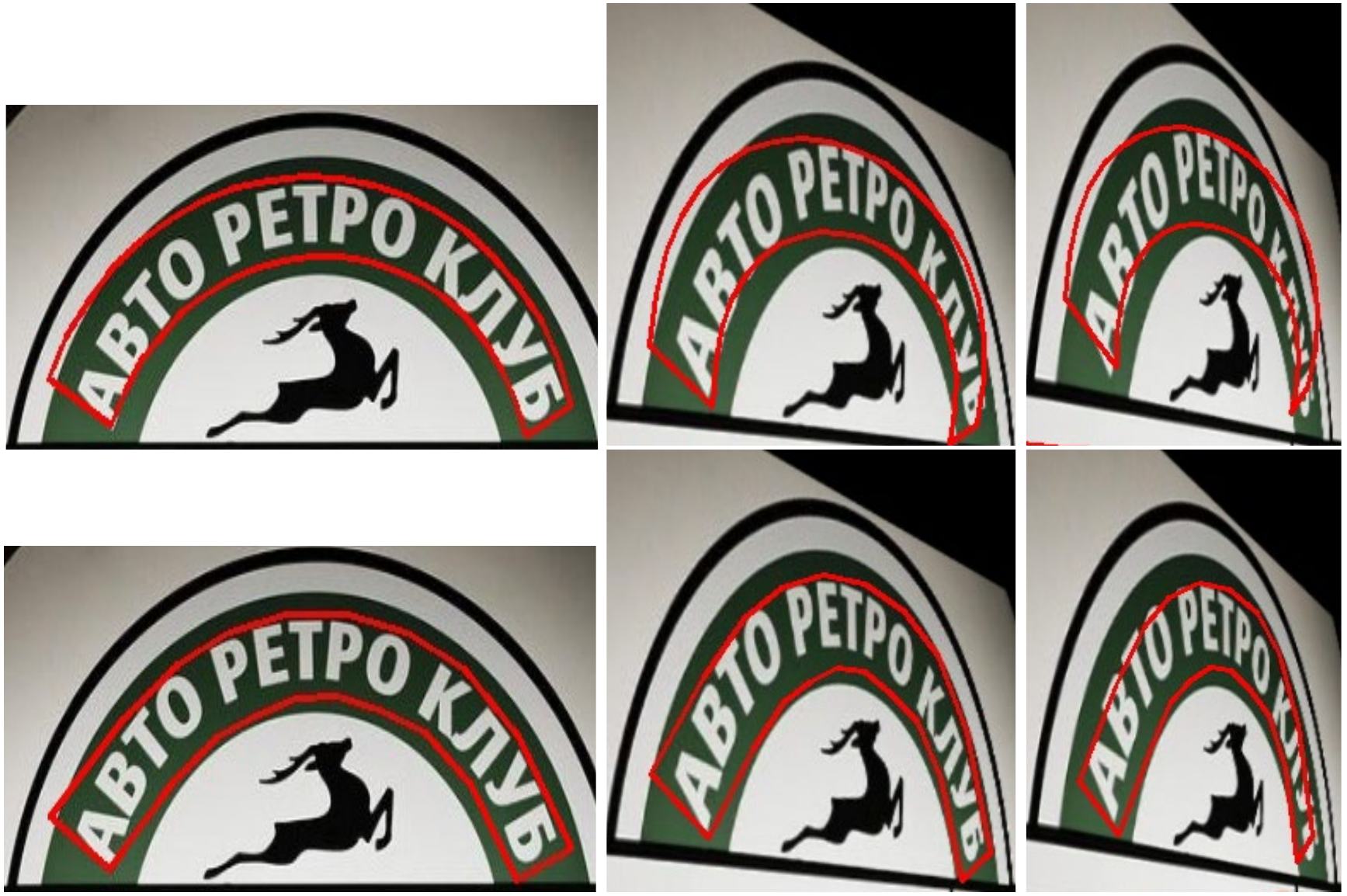}
		\caption{Visualization of the detection results of Bezier representation (top) and TPS representation (bottom) on highly curved text. From left to right, the image is transformed to larger degrees of perspective. }
		\label{fig:perspective}
		\vspace{-15px}
	\end{figure}
	
	\subsubsection{Compared with Bezier}
	Bezier is another impressive representation for the arbitrary shape text, and it employs a  different formula from TPS. To compare with Bezier fairly, we replace the TPS representation with the Bezier representation of 8 control points in our framework, and both models of TPS and Bezier are trained with the same loss functions on the training set of CTW1500. Following  \cite{zhu2021fourier}, the highly curved are sampled from the test set of CTW1500. Nearly all the images in the dataset are focused on the text with no perspective, but in the real world, scene texts are often viewed sideways. To simulate this situation, we apply a perspective transformation to the images: Rotate the image along the left edge and then reproject the image onto the plane to get the perspective image. The perspective transformation can be applied as a data augmentation to the training process, where the rotation angle is randomly selected from the range of $0^\circ$- $70^\circ$, and the highly curved test subset is transformed with the rotation angle at $0^\circ$, $45^\circ$ and $70^\circ$ to evaluate the TPS and Bezier representation. The results are shown in Table~\ref{tab:bezier_vs_tps}. Without perspective augmentation, our TPSNet performs slightly better than Bezier. With perspective augmentation, our TPSNet is improved much more than Bezier, especially on a perspective angle at $70^\circ$ (9.0 vs. 5.3). The visualization results are shown in Fig.~\ref{fig:perspective}. As discussed in Section~3.1, our TPS representation has more flexibility to fit the highly curved text with perspective, while the Bezier with limited control points can not. The results prove that the TPS better represents the arbitrary shape scene text.

	\subsubsection{Evaluation with TIOU Metric}
\begin{figure}[t]
		\vspace{-0.15cm} 
		\centering
		\setlength{\abovecaptionskip}{5px}
		\subfigbottomskip=-3pt
		\subfigcapskip=-5pt
		\subfigure[TextRay \cite{Wang2020textray}]{\includegraphics[width=0.49\linewidth]{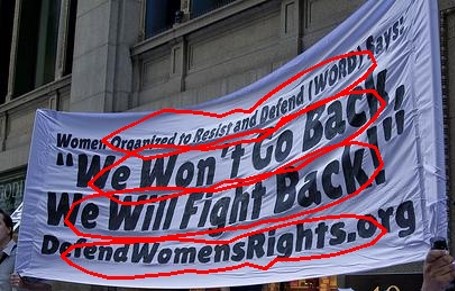}}\hspace{1.5pt}
		\subfigure[FCENet \cite{zhu2021fourier}]{\includegraphics[width=0.49\linewidth]{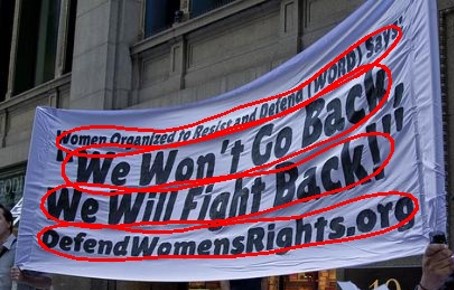}}\vspace{-2pt}\\
		\subfigure[BPN \cite{zhang2021adaptive}]{\includegraphics[width=0.49\linewidth]{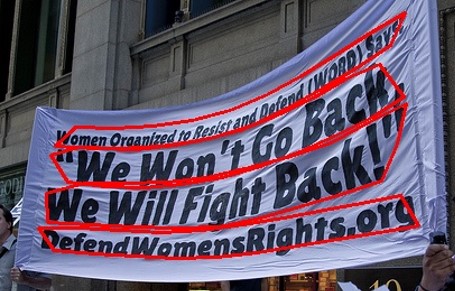}}\hspace{1.5pt}
		\subfigure[PCR \cite{dai2021progressive}]{\includegraphics[width=0.49\linewidth]{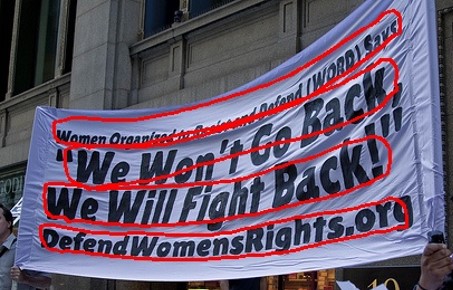}}\vspace{-2pt}\\
		\subfigure[ABCNetV2 \cite{abcnetv2}]{\includegraphics[width=0.49\linewidth]{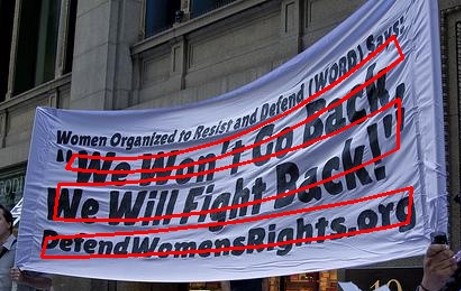}}\hspace{1.5pt}
		\subfigure[TPSNet (Ours)]{\includegraphics[width=0.49\linewidth]{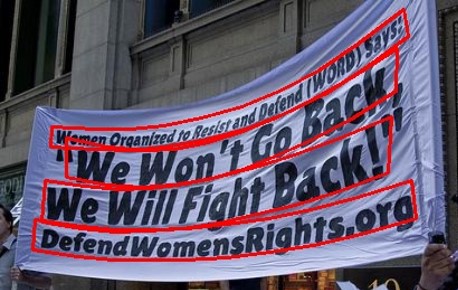}}
		\caption{Qualitative comparisons with previous methods on selected challenging samples in CTW1500.}
		\label{fig:quali}
		\vspace{-13px}
	\end{figure}

	TIOU \cite{liu2019tightness} is a well-known evaluation protocol for scene text detection. TIOU-Recall and TIOU-Precision can separately quantify the completeness and compactness of the detection results, and the TIOU-Hmean quantifies the overall tightness of the matching degree. Tighter detection means less miss of character and less presence of background, which is a stricter metric than the VOC metric. As shown in Table \ref{tab:tiou}, our TPSNet achieves the best performance on all three metrics. The FCENet \cite{zhu2021fourier} is the most competitive method with our method on the common VOC metric, but on the TIOU metric, TPSNet outperforms it by 3.6\%, which means that the detection results from TPSNet are tighter than previous methods.
	Qualitative comparison is shown in Fig. \ref{fig:quali}. ABCNetV2 \cite{Liu2020ABCNet} and PCR \cite{dai2021progressive} fail in long and dense texts, while TextRay \cite{Wang2020textray}, FCENet \cite{zhu2021fourier} and BPN \cite{zhang2021adaptive} prefer to missing the corners of long text, which is not conducive to the subsequent recognition. By comparison, our proposed TPSNet obtained the most compact and complete detection.
	
	\begin{table}[t]
		\setlength{\abovecaptionskip}{0cm}  %段前
		\small
		\caption{Evaluation with TIOU \cite{liu2019tightness} metric on CTW1500. Detection-only training is applied.}
		\centering
		\begin{tabularx}{\linewidth}{@{}lYYY@{}}
			\toprule
			Methods & TIOU-R & TIOU-P & TIOU-H \\ \midrule
			FCENet \cite{zhu2021fourier}  & 58.6   & 69.8   & 63.7   \\
			PCR \cite{dai2021progressive}     & 58.0     & 69.9   & 63.4   \\
			BPN \cite{zhang2021adaptive} & 61.5   & 69.2   & 65.1   \\ \midrule
			TPSNet(ours)  & \textbf{62.9}   & \textbf{72.4}   & \textbf{67.3}   \\ \bottomrule
		\end{tabularx}
		\label{tab:tiou}
		\vspace{-15px}
	\end{table}
	
	\begin{table*}[ht]
		\setlength{\abovecaptionskip}{0cm}  %段前
		\small
		\caption{Comparison with previous detection methods on CTW1500, Total-Text and ICDAR2015 test sets. ‘Ext’ means using the external dataset to pretrain the model. * denotes the results based on end-to-end text spotting training.}
		\centering
		\renewcommand{\arraystretch}{0.8}
		\begin{tabularx}{\linewidth}{@{}lYYYYYYYYYYYYYX@{}}
			\toprule
			\multicolumn{1}{l}{\multirow{2}{*}{Methods}} &
			\multicolumn{1}{c}{\multirow{2}{*}{Ext}} &
			\multicolumn{4}{c}{CTW1500} &
			\multicolumn{4}{c}{Total-Text} &
			\multicolumn{4}{c}{ICDAR2015} \\ \cmidrule(l){3-14} 
			\multicolumn{1}{c}{} & \multicolumn{1}{c}{} & R    & P    & H    & FPS  & R    & P    & H    & FPS  & R    & P    & H    & FPS  \\ \midrule
			TextSnake \cite{long2018textsnake}            & \checkmark                 & \textbf{85.3} & 67.9 & 75.6 & $-$ & 74.5 & 82.7 & 78.4 & $-$ & 80.4 & 84.9 & 82.6 & 1.1  \\
			PSENet \cite{wang2019PSENet}               & \checkmark                 & 79.7 & 84.8 & 82.2 & 3.9  & 78.0 & 84.0 & 80.9 & 3.9  & 84.5 & 86.9 & 85.7 & 1.6  \\
			CRAFT \cite{baek2019CRAFT}                & \checkmark                 & 81.1 & 86.0 & 83.5 & $-$ & 79.9 & 87.6 & 83.6 & $-$ & 84.3 & 89.8 & 86.9 & 8.6  \\
			SAE \cite{tian2019SAE}                  & \checkmark                 & 77.8 & 82.7 & 80.1 & $-$ & 77.8 & 82.7 & 80.1 & $-$ & 85.0 & 88.3 & 86.6 & $-$ \\
			MSR \cite{xue2019msr}                  & \checkmark                 & 78.3 & 85.0 & 81.5 & 4.3  & 74.8 & 83.8 & 79.0 & 4.3  & 78.4 & 86.6 & 82.3 & $-$ \\
			PAN \cite{wang2019PAN}                  & \checkmark                 & 81.2 & 86.4 & 83.7 & \textbf{39.8} & 81.0 & 89.3 & 85.0 & \textbf{39.6} & 81.9 & 84.0 & 82.9 & \textbf{26.1} \\
			SAST \cite{wang2019SAST}                 & \checkmark                 & 77.1 & 85.3 & 81.0 & $-$ & 76.9 & 83.8 & 80.2 & $-$ & 87.1 & 86.7 & 86.9 & $-$ \\
			%TextField \cite{xu2019textfield}            & \checkmark                 & 79.8 & 83.0 & 81.4 & 6.0  & 79.9 & 81.2 & 80.6 & 6.0  & 83.9 & 84.3 & 84.1 & 6    \\
			DB \cite{liao2020db}                   & \checkmark                 & 80.2 & 86.9 & 83.4 & 22.0 & 82.5 & 87.1 & 84.7 & 32.0 & 83.2 & \textbf{91.8} & 87.3 & 12   \\
			DRRGN \cite{zhang2020DRRG}                & \checkmark                 & 83.0 & 85.9 & 84.5 & $-$ & 84.9 & 86.5 & 85.7 & $-$ & 84.7 & 88.5 & 86.6 & 3.5  \\
			%CRNet \cite{zhou2020crnet}                & \checkmark                 & 80.9 & 87.0 & 83.8 & $-$ & 82.5 & 85.8 & 84.1 & $-$ & 84.5 & 88.3 & 86.4 & $-$ \\ 
			\midrule
			%SPCNet \cite{xie2019SPCNet}               & \checkmark                 & $-$ & $-$ & $-$ & $-$ & 82.8 & 83.0 & 82.9 & $-$ & 85.8 & 88.7 & 87.2 & $-$ \\
			LOMO \cite{zhang2019lomo}                 & \checkmark                 & 69.6 & \textbf{89.2} & 78.4 & 4.4  & 75.7 & 88.6 & 81.6 & 4.4  & 83.5 & 91.3 & 87.2 & 3.4  \\
			CSE \cite{liu2019CSE}                  & $\times$                 & 76.0 & 81.1 & 78.4 & 0.4  & 79.1 & 81.4 & 80.2 & 0.4  & $-$ & $-$ & $-$ & $-$ \\
			ContourNet \cite{Wang2020contournet}           & $\times$                 & 84.1 & 83.7 & 83.9 & 4.5  & 83.9 & 86.9 & 85.4 & 3.8  & 86.1 & 87.6 & 86.9 & 3.5  \\
			Mask-TTD \cite{liu2019maskTTD}             & $\times$                 & 79.0 & 79.7 & 79.4 & $-$ & 74.5 & 79.1 & 76.7 & $-$ & 87.6 & 86.6 & 87.1 & $-$ \\
			MaskTextSpotter* \cite{liao2019masktextspotterv2}                        & \checkmark           & $-$ & $-$ & $-$ & $-$ & 82.4 & 88.3 & 85.2 & $-$ & 87.3 & 86.6 & 87.0 & $-$ \\ \midrule
			ATRR \cite{wang2019ATRR}                 & $\times$                 & 80.2 & 80.1 & 80.1 & 10.0 & 76.2 & 80.9 & 78.5 & 10.0 & 86.0 & 89.2 & 87.6 & 10.0   \\
			%Boundary             & \checkmark                 & $-$ & $-$ & $-$ & $-$ & 83.5 & 85.2 & 84.3 & $-$ & 88.1 & 82.2 & 85.0 & $-$ \\
			Boundary*  \cite{wang2019boundary}           & \checkmark                 & $-$ & $-$ & $-$ & $-$ & 85.0 & 88.9 & 87.0 & $-$ & 87.5 & 89.8 & 88.6 & $-$ \\
			%ABCNet               & \checkmark                 & 78.5 & 84.4 & 81.4 & $-$ & 81.3 & 87.9 & 84.5 & $-$ & $-$ & $-$ & $-$ & $-$ \\
			TextRay \cite{Wang2020textray}              & $\times$                 & 80.4 & 82.8 & 81.6 & $-$ & 77.9 & 83.5 & 80.6 & $-$ & $-$ & $-$ & $-$ & $-$ \\
			%FCENet               & $\times$                 & 80.7 & 85.7 & 83.1 & $-$ & 79.8 & 87.4 & 83.4 & $-$ & $-$ & $-$ & $-$ & $-$ \\
			FCENet \cite{zhu2021fourier}               & $\times$                 & 83.4 & 87.6 & 85.5 & $-$ & 82.5 & 89.3 & 85.8 & $-$ & 82.6 & 90.1 & 86.2 & $-$ \\
			%PCR                  & $\times$                 & 79.8 & 85.3 & 82.4 & $-$ & 80.2 & 86.1 & 83.1 & $-$ & $-$ & $-$ & $-$ & $-$ \\
			PCR \cite{dai2021progressive}                  & \checkmark                 & 82.3 & 87.2 & 84.7 & 11.8 & 82.0 & 88.5 & 85.2 & $-$ & $-$ & $-$ & $-$ & $-$ \\
			MOST \cite{he2021most}                 & \checkmark                 & $-$ & $-$ & $-$ & $-$ & $-$ & $-$ & $-$ & $-$ & \textbf{89.1} & 87.3 & 88.2 & 10.0 \\
			BPN \cite{zhang2021adaptive}              & \checkmark                 & 83.6 & 86.5 & 85.0 & 12.2 & 85.2 & \textbf{90.7} & 87.9 & 10.7 & $-$ & $-$ & $-$ & $-$ \\ 
			ABCNetV2* \cite{abcnetv2}            & \checkmark                 & 83.8 & 85.6 & 84.7 & $-$ & 84.1 & 90.2 & 87.0 & $-$ & 86.0 & 90.4 & 88.1 & $-$ \\ \midrule
			TPSNet(ours)         & $\times$                 & 83.7 & 88.1 & 85.9 &  17.9    & 84.0 & 89.2 & 86.6 &   14.3   & 85.1 & 90.5 & 87.7 &  11.6    \\
			TPSNet(ours)         & \checkmark                 & 85.1 & 87.7 & 86.4 &17.9      & 86.8 & 89.5 & 88.1 & 14.3     & 86.6 & 90.7 & 88.6 &   11.6   \\ 
			TPSNet(ours)*         & \checkmark           &  86.3    &  88.7    &    \textbf{87.5}  & 17.9 &   \textbf{86.8}   & 90.2     & \textbf{ 88.5}    & 14.3 & 87.8     &  90.5    &  \textbf{89.1}    & 11.6 \\ \bottomrule
		\end{tabularx}
		\label{table:sota}
		\vspace{-13px}
	\end{table*}
	
	\begin{table}[t]
		\setlength{\abovecaptionskip}{0cm}  %段前
		\small
		\caption{Comparison with previous methods on ArT. ‘Ext’means using the external dataset to pretrain the model.}
		\centering
		\renewcommand{\arraystretch}{0.8}
		\begin{tabularx}{\linewidth}{@{}lYYYYX@{}}
			\toprule
			Method       & Ext  & R    & P    & H    \\ \midrule
			TextRay \cite{Wang2020textray}      & \checkmark & 58.6 & 76.0 & 66.2 \\
			PCR \cite{dai2021progressive}          & \checkmark & 66.1 & 84.0 & 74.0 \\ \midrule
			TPSNet(Ours) & $\times$ & 70.9 & 81.0 & 75.6 \\
			TPSNet(Ours) & \checkmark & \textbf{73.3} & \textbf{84.3} & \textbf{78.4} \\ \bottomrule
		\end{tabularx}
		\label{tab:art}
		\vspace{-10px}
	\end{table}
	
	\subsubsection{Evaluation with Benchmark Metric}
	We evaluate our TPSNet on benchmark datasets and compare with previous detection methods as shown in Table \ref{table:sota} and \ref{tab:art}. Previous methods are divided into three categories: segmentation-based, regression-based, and hybrid-based methods that use both segmentation and regression for detection.  As a regression-based method, our TPSNet can outperform previous methods on curved scene text detection datasets CTW1500 and Total-Text.
	For multi-oriented scene text in ICDAR2015, our proposed TPSNet can also achieve the best performance because the TPS representation adopts the rectangular fiducial shape, and when it is applied for multi-oriented text, the TPS parameters $\mathbf{T}$ will collapse to $[c_x , a_{1_x} , a_{2_x}; c_x , a_{1_y} , a_{2_y}]$ that denotes an affine transformation that is still effective.
	%it can also achieve the best performance. 
	In addition, even without extra data for pretraining, the TPSNet is still comparable with or even better than previous methods with extra data, which proves that the TPSNet is easy to train.  Thanks to the simplicity and efficiency of the TPS representation, TPSNet also has real-time inference speed.
	
	We also evaluate our model on Art, which is the most challenging arbitrary shape scene text detection dataset. as shown in Table \ref{tab:art}, our TPSNet can boost the Hmean from 74.0\% to 78.4\% compared with previous best regression-based method PCR \cite{dai2021progressive}. The 4.4\% improvement proves that TPS representation is adaptive for various practical situations.

	\subsection{
		%Comparisons with Previous Text Spotters
		End-to-end Evaluation}
	Concatenated with a simple recognition module, we extend our TPSNet to an end-to-end text spotter with ease. The recognition module and end-to-end training strategy are both following ABCNet V2 \cite{abcnetv2}. As shown in Table \ref{tab:e2e}. Our TPSNet achieves 76.1\% on Total-Text and 59.7\% on CTW1500 in end-to-end Hmean without lexicon, which outperforms all previous text spotters. Note that our TPSNet does not need any character-level annotations to supervise the network like MaskTextSpotter \cite{liao2020masktextspotterv3} or MANGO \cite{qiao2021mango}. The performance of MANGO \cite{qiao2021mango} is closest to ours, but its predicted pixel-level character classification slows down its inference speed.

	\begin{table}[t]
		\setlength{\abovecaptionskip}{0cm}  %段前
		\small
		\caption{Text spotting performance on Total-Text and CTW1500 datasets. “None” means lexicon-free, and “Full” represents using all the words appeared in the test set as lexicon. * denotes results based on multi-scale test. }
		\centering
		\renewcommand{\arraystretch}{0.8}
		\begin{tabularx}{\linewidth}{@{}lYYYYYX@{}}
			\toprule
			\multirow{2}{*}{Methods} & \multicolumn{3}{c}{Total-Text} & \multicolumn{2}{c}{CTW1500} \\ \cmidrule(l){2-6} 
			& None     & Full     & FPS      & None       & Full       \\ \midrule
			%TextBoxes++      \cite{liao2018textboxes++}         & 36.3     & 48.9     & 1.4      & $-$        & $-$        \\
			%FOTS             \cite{liu2018FOTS}         & $-$      & $-$      & $-$      & 21.1       & 39.7       \\
			%MaskTextSpotterv1  \cite{lyu2018masktextspotter}       & 52.9     & 71.8     & 4.8      & $-$        & $-$        \\
			%MaskTextSpotterv2 \cite{liao2019masktextspotterv2}        & 65.3     & 77.4     & 2.0      & $-$        & $-$        \\
			TextDragon       \cite{feng2019textdragon}         & 44.8     & 74.8     & 2.6      & 39.7       & 72.4       \\
			Unconstrained \cite{qin2019towards}           & 67.8     & $-$      & 4.8      & $-$        & $-$        \\
			CharNet          \cite{xing2019CCN}         & 66.6     & $-$      & 1.2      & $-$        & $-$        \\
			ABCNet           \cite{Liu2020ABCNet}         & 64.2     & 75.7     & 17.9     & 45.2       & 74.1       \\
			Boundary         \cite{wang2019boundary}         & 65.0     & 76.1     & $-$      & $-$        & $-$        \\
			TextPerceptron   \cite{qiao2020textperceptron}         & 69.7     & 78.3     & $-$      & 57.0       & $-$        \\
			PGNet            \cite{wang2021pgnet}         & 60.5     & $-$      & \textbf{40.5}     & $-$        & $-$        \\
			MANGO               \cite{qiao2021mango}       & 72.9     & 83.6     & 4.3      & 58.9       & 78.7       \\
			MaskTextSpotterv3   \cite{liao2020masktextspotterv3}      & 71.2     & 78.4     & 2.5      & $-$        & $-$        \\
			PAN++            \cite{wang2021pan++}            & 68.6     & 78.6     & 21.1     & $-$        & $-$        \\
			ABCNetV2          \cite{abcnetv2}                & 70.4     & 78.1     & 10.0     & 57.5       & 77.2       \\
			ABCNetV2*          \cite{abcnetv2}                & 73.5     & 80.7     & $-$     & 58.4       & 79.0       \\
			%Feng et al \cite{feng2021residual}               & 55.8     & 79.2     & 7.2      & 42.2       & 74.9       \\ 
			\midrule
			%TPSNet w/o e2e-optim                            & 73.8     & 79.6   &  9.3       & 58.8            & 78.3                \\
			TPSNet(Ours)                      & 76.1     & 82.3     & 9.3         & 59.7       & 79.2      \\ 
			TPSNet(Ours)*                      & \textbf{78.5}     & \textbf{84.1}     & $-$         & \textbf{60.5}       & \textbf{80.1}\\
			\bottomrule
		\end{tabularx}
		\label{tab:e2e}
		\vspace{-15px}
	\end{table}
	
	\section{Conclusion}
	In this paper, we propose a novel TPS representation for arbitrary shape text, which comes from the sophisticated reverse thinking of Thin Plate Splines. The TPS representation is compact, complete, efficient, and reusable for subsequent recognition. To further exploit the potential of the TPS representation, the Border Alignment Loss is designed. Based on the representation and loss function, we implement the TPSNet and extend it to a text spotter. The TPSNet is evaluated on CTW1500, Total-Text, ICDAR2015, and Art datasets for text detection and spotting, and it outperforms previous counterparts with large margins.
	\begin{acks}
		Supported by the Beijing Municipal Science \& Technology Commission (Z191100007119002), the Key Research Program of Frontier Sciences, CAS, Grant NO ZDBS-LY-7024.
	\end{acks}
	
	\bibliographystyle{ACM-Reference-Format}
	\balance
	\bibliography{ref}

	\newpage
	\appendix
	\section{TPS Representation}
	\subsection{Evaluation about fitting abilities}
	\label{sec:a1}
	To verify the fitting ability of TPS,
	we solve the TPS parameters $\mathbf{T}$ with the standard least square method from equation (\ref{equ:5}).
	The fitting boundary can be derived with equation (\ref{equ:1}), then evaluated the 
	Tightness-IOU (TIOU) \cite{liu2019tightness} comparing with the ground truth annotations. The TIOU value reveals the compactness and completeness of the fitting.
	The fitting results of different representations in several previous methods are demonstrated in Table \ref{tab:fittingmeasure}, where TextRay and Fourier use 22-dimensional parameters, and both Bezier and TPS use 8 control points.
	\begin{table}[h]
		% \scriptsize   
		%\setlength\tabcolsep{3pt}
		\setlength{\abovecaptionskip}{0cm}  %段前
		\small
		\caption{Comparison of text shape representation. TIOU  \cite{liu2019tightness} means the tightness-IOU between the fitted shapes and ground truth. ``Corner", ``Rectify" means whether the representation can keep the corner, can rectify irregular text for recognition.}
		\centering
		\renewcommand{\arraystretch}{0.8}
		\begin{tabularx}{\linewidth}{lYYY}
			\toprule
			Method    & TIOU-H     & Corner & Rectify   \\ \midrule
			TextRay \cite{Wang2020textray}   & 82.5      & $\times$    & $\times$     \\
			Fourier \cite{zhu2021fourier}   & 90.6       & $\times$    & $\times$     \\
			Bezier \cite{Liu2020ABCNet, abcnetv2}     & 96.8     & \checkmark   & \checkmark    \\
			TPS (Ours) & 97.1      & \checkmark   & \checkmark    \\ \bottomrule
		\end{tabularx}
		\label{tab:fittingmeasure}
		\vspace{-12px}
	\end{table}
	
	% \section{Distributions of Fiducial Points}
	\subsection{Distributions of Fiducial Points}
	\label{sec:a2}
	The fiducial points with three different distributions can define different basis functions $\phi(x,y)$. To illustrate their differences, we use the standard least square method to solve the parameters for Bezier and TPS to fit the polygon annotations, and the results are shown in Figure~\ref{fig:dist}. For Bezier, the solved parameters are the coordinates of the control points, and for TPS, the parameter $T$ is solved. As the blue points shown in (b), (c), and (d), with $T$, the fiducial points can be transformed into corresponding points on the text shape, and they will locate at the similar relative positions as the fiducial points, which can be seen as the control points for TPS. Obviously, with the same number of control points, Bezier fails to fit this highly curved text, but TPS works well. Furthermore, the "Cross" distribution can achieve the best shape fitting and rectification results. With the "Cross" distribution, the fiducial points alternately distribute at the edge and center of the text, balancing the length of the text and the accuracy of the edges.
	% \subsection{Ablation Study}
	\begin{table}[h]
		\setlength{\abovecaptionskip}{0cm}  %段前
		\small
		\caption{Ablation study about different distribution of the fiducial points on CTW1500. (b), (c) and (d) refer to the distributions in Fig.~\ref{fig:fid}.}
		\centering
		\begin{tabularx}{\linewidth}{@{}lYY@{}}
			\toprule
			Fiducial Distribution & H(IOU@0.5) & H(IOU@0.7) \\ \midrule
			(b) Edge            &  85.5              &   74.3             \\
			(c) Cross          &  \textbf{85.9}              &   \textbf{76.0}             \\
			(d) Center           &  85.6              &   75.1             \\ \bottomrule
		\end{tabularx}
		\label{tab:fiducial}
		\vspace{-5px}
	\end{table}
	
	For the distribution of the fiducial points, we conduct the ablation study on CTW1500 without pretraining. 
	As shown in Table~\ref{tab:fiducial}, three distributions of the fiducial points in Fig.~\ref{fig:fid} are applied to construct the TPS representation. The performance is evaluated  with IOU constraints at 0.5 and 0.7.
	When evaluated with IOU@0.5, there are only slight differences between these three distributions. But when evaluated with IOU@0.7, the second distribution $Cross$ is clearly better than the others, indicating the crossed distribution of fiducial points at the edge and center allows for the better shape representation of the TPS.
	
	\begin{figure}[h]
		\subfigbottomskip=-3pt
		\subfigcapskip=-5pt
		\setlength{\abovecaptionskip}{3px}
		\setlength{\belowcaptionskip}{-0.2cm}
		\centering
		\subfigure[Bezier]{\includegraphics[width=0.49\linewidth]{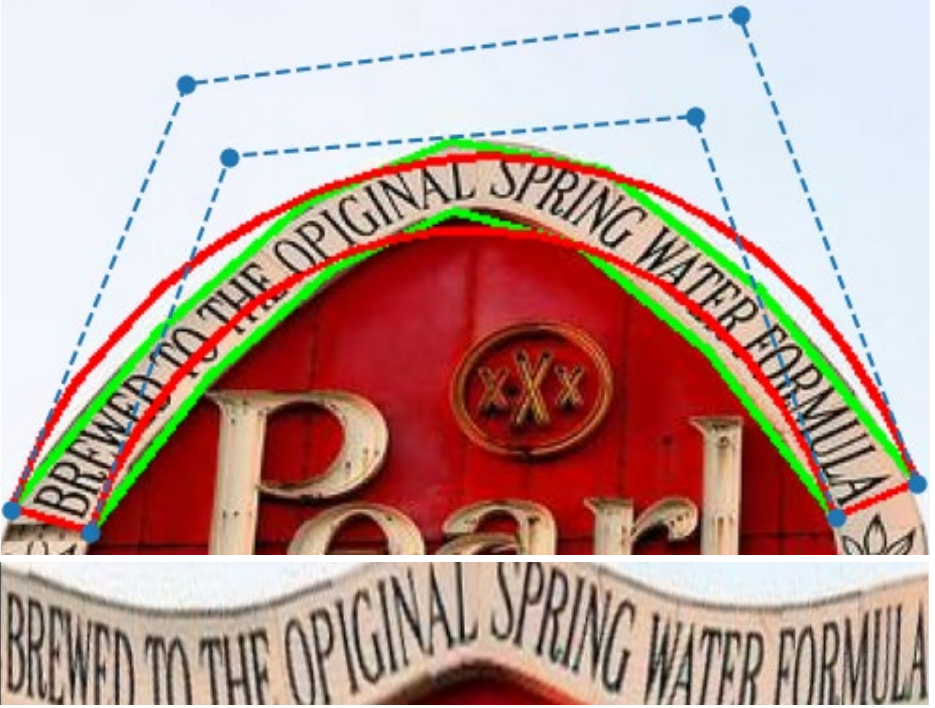}}
		\subfigure[Edge]{\includegraphics[width=0.49\linewidth]{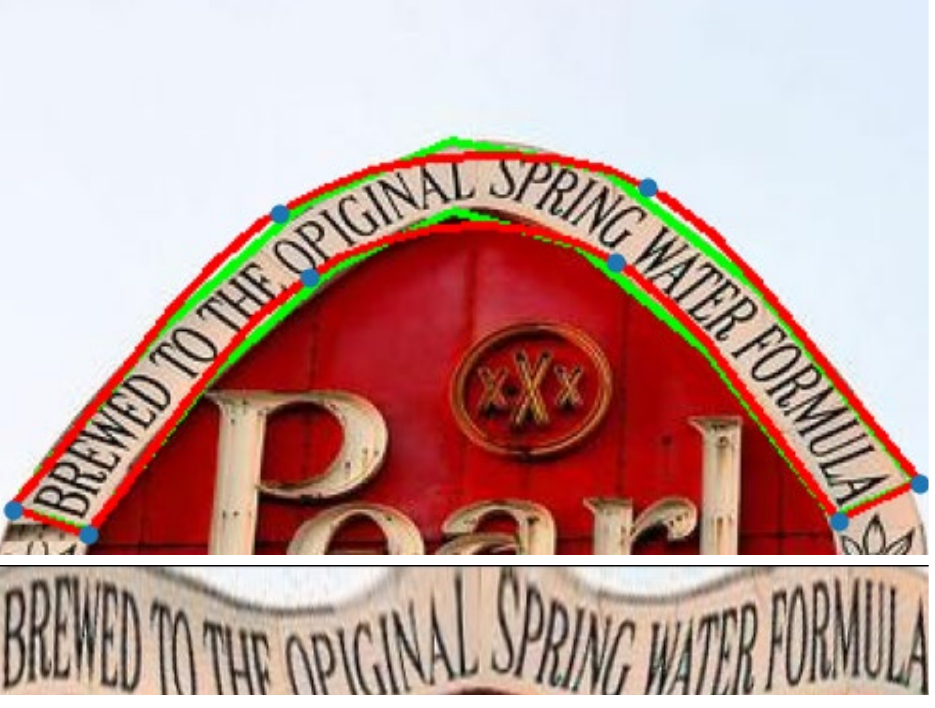}}
		\subfigure[Cross]{\includegraphics[width=0.49\linewidth]{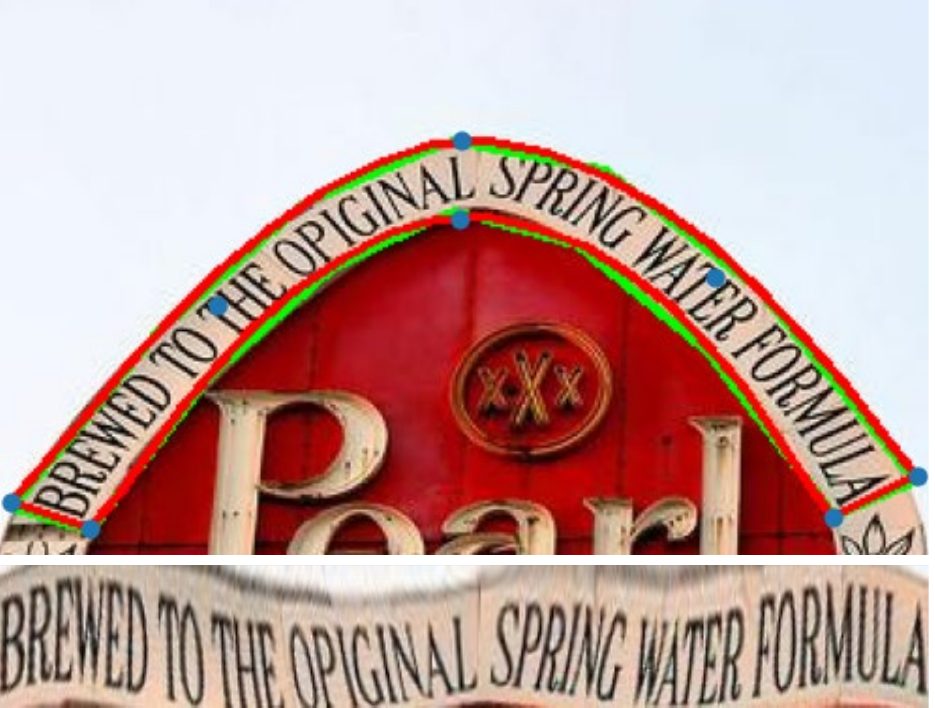}}
		\subfigure[Center]{\includegraphics[width=0.49\linewidth]{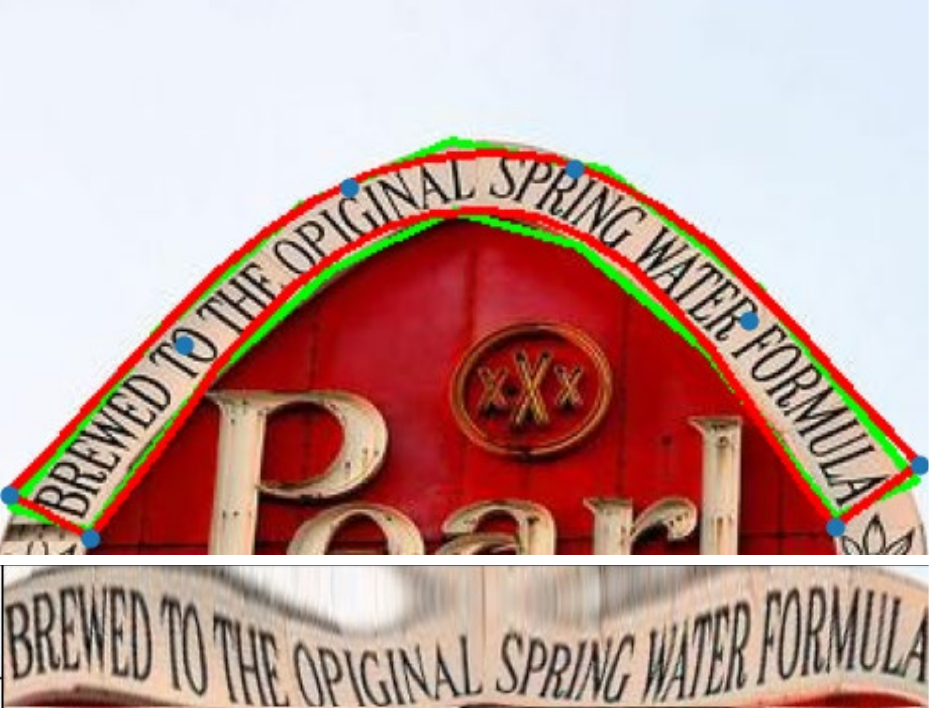}}
		\caption{Visualizations of fitting results on the highly curved text with Bezier (a) and TPS defined by three different distributions of fiducial points. The green polygons are the ground truth, red lines are fitting results, and blue points are control points. The bottom of the image is the result of the rectification.}
		\label{fig:dist}
	\end{figure}
	\section{TPSNet}
	\subsection{Relaxation Threshold}
	\label{sec:b1}
	The ablation study about the border relaxation threshold $t_r$ is conducted on Total-Text, and the results are shown in Table~\ref{tab:tr}. $t_r = 1.0$ means no relaxation is applied to the border mask ground truth, and smaller $t_r$ leads to larger relaxation range. From the experiments, the border relaxation with $t_r = 0.8$ achieves the best performance, and more relaxation is harm for the model optimization.
	\begin{table}[h]
		\setlength{\abovecaptionskip}{0cm}  %段前
		\centering
		\caption{The ablation study about the relax threshold $t_r$ on Total-Text. Detection-only training is applied without pretraining.}
		\renewcommand{\arraystretch}{0.8}
		\small
		\begin{tabularx}{\linewidth}{@{}lYYY@{}}
			\toprule
			$t_r$ & R    & P    & H    \\ \midrule
			1.0   & 84.6 & 87.1 & 85.8 \\
			0.9   & \textbf{84.6} & 88.1 & 86.3 \\
			0.8   & 84.0 & \textbf{89.2} & \textbf{86.6} \\
			0.7   & 83.6 & 87.4 & 85.4 \\ \bottomrule
		\end{tabularx}
		\label{tab:tr}
	\end{table}
	
	\begin{figure*}[t]
		\centering
		\includegraphics[width=0.95\linewidth]{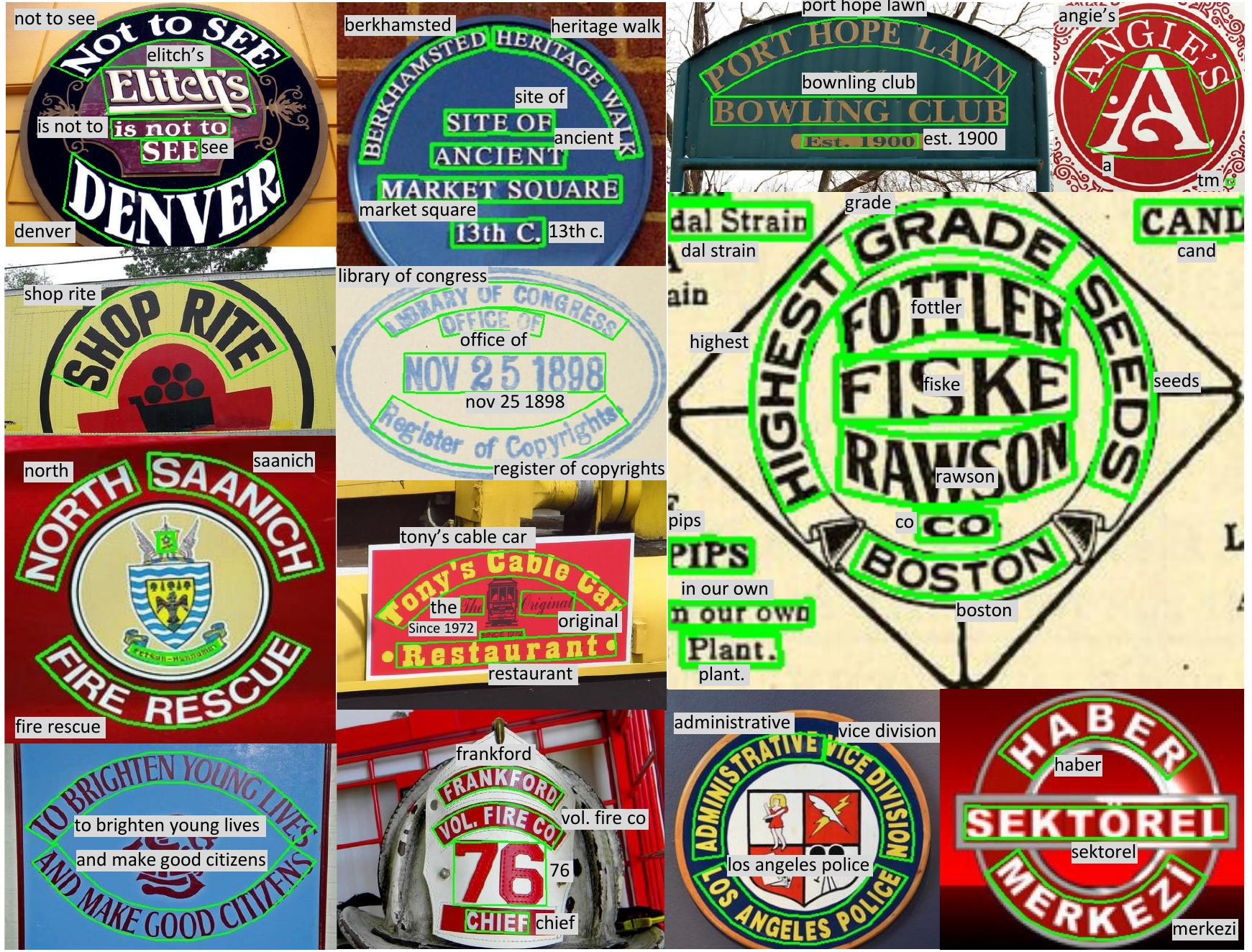}
		\caption{Visualizations of spotting results on test samples of CTW1500. Green lines are detection results, and the recognition results are marked nearby.}
		\label{fig:visctw}
	\end{figure*}
	
	\begin{figure}
		\centering
		\setlength{\abovecaptionskip}{5px}
		\subfigbottomskip=-3pt
		\subfigcapskip=-5pt
		% \subfigure[]{\includegraphics[width=0.48\linewidth]{figures/img144_text_region.png}}
		\subfigure[Text Center Region]{\includegraphics[width=0.48\linewidth]{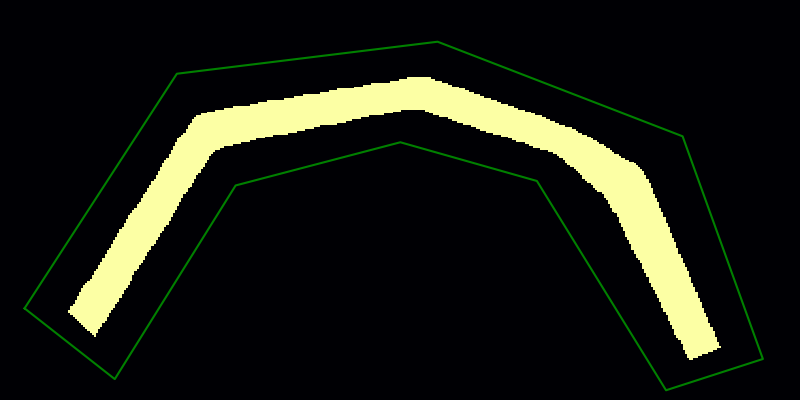}}
		% \subfigure[Gaussian Text Center]{\includegraphics[width=\linewidth]{figures/gauss_c.pdf}}
		% \subfigure[Gaussian Text Center]{\includegraphics[width=0.48\linewidth]{img144_gauss_center.png}}
		\vspace{2pt}
		\subfigure[Illustration of the generation process of Guassian Text Center.]{\includegraphics[width=\linewidth]{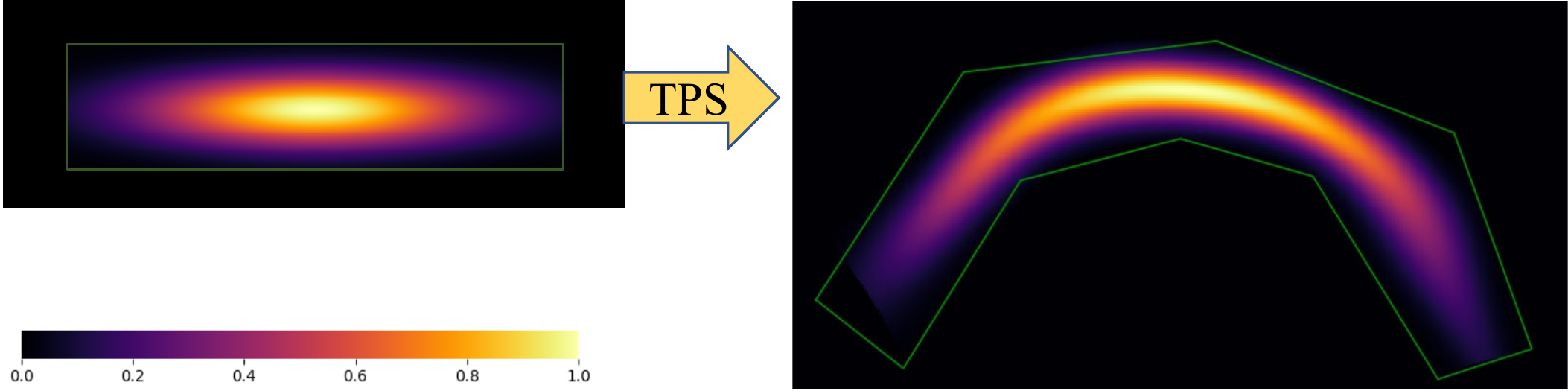}}
		\caption{Illustration of the Text Center \cite{long2018textsnake,zhu2021fourier} and our proposed Gaussian Text Center.}
		\label{fig:gauss_center}
		\vspace{-10px}
	\end{figure}
	
	\subsection{Gaussian Text Center}
	\label{sec:b2}
	In the previous work FCENet \cite{zhu2021fourier}, TC is the shrink region from the text region as shown in Fig.~\ref{fig:gauss_center}~(a), which is nearly as long as the text. However, this design is not suitable for a one-stage regression-based model because the position near one end of the long text can hardly perceive the shape of the other end, so only the prediction around the center point should be reserved. 
	Centerness \cite{Tian2019FCOS} is designed for the general object detection to solve this problem, but it does not suit for the arbitrary shape text. To this end, we propose the Gaussian Text Center as shown in Fig.~\ref{fig:gauss_center}~(b). 
	% Its ground truth is generated as follows: generate a two-dimensional gaussian distribution as the centerness for a rectangle, and then the distribution map is transformed to the arbitrary shape of the text with the TPS transformation according to the boundary annotation.
	The ground truth of the Gaussian Text Center (GTC) is generated as the process shown in Figure~\ref{fig:gauss_center}~(b). Firstly, we generate a two-dimensional gaussian distribution map in a rectangle, then we use TPS transformation to transform the rectangle to the arbitrary text shape, which is exactly the reverse process of text rectification. Note that, the center map does not need high precision, so using the noise annotation to solve this transformation will not affect the performance. The mask value on GTC varies continuously in range $0-1$, so the classification loss for the text center $L_{TC}$ becomes soft cross entropy loss:
	\begin{equation}
	L_{TC} =- \sum_i^N y_i \log p_i + (1-y_i) \log{(1-p_i)}
	\end{equation}
	where $y_i$ is the value of $i_{th}$ pixel on GTC map, and $p_i$ is the predicted confidence.
	
	% \begin{figure}
	%     \centering
	%     % \setlength{\belowcaptionskip}{-0.5cm}
	%     \subfigbottomskip=-3pt
	%     \subfigcapskip=-5pt
	%     \includegraphics[width=\linewidth]{figures/gauss_c.pdf}
	%     \caption{Illustration of the generation process of Guassian Text Center.}
	%     \label{fig:gauss}
	% \end{figure}
	\section{Visualizations of Spotting Results}
	Visualizations of spotting results on test samples of CTW1500 are shown in Figure~\ref{fig:visctw}.  For curved texts, our TPSNet can detect the shape of the text completely and compactly and  identify the text content with the rectification accurately.

\end{document}